\documentclass[journal]{IEEEtran}

\usepackage{graphicx}
\usepackage{booktabs}
\usepackage{array}
\usepackage{graphics}
\usepackage{multirow}
\usepackage{utfsym}
\usepackage{graphicx}
\usepackage{epstopdf}
\usepackage{upgreek}
\usepackage{amsmath}
\usepackage{amsfonts}
\usepackage{subcaption}
\usepackage{tabularray}
\usepackage[numbers,sort&compress]{natbib}
\usepackage[colorlinks,linkcolor=black]{hyperref}
\usepackage{bm}
\usepackage{cancel}
\usepackage{algorithm}
\usepackage{algpseudocode}

\newcommand{\dt}[1]{\iffalse{#1}\fi}

\newcommand{\dd}[1]{{\color{black}{#1}}}
\usepackage{xcolor}
\newcommand{\bb}[1]{\textcolor{black}{#1}}

\usepackage{amssymb}
\usepackage{pifont}

\newcommand{\cmark}{\ding{51}}  
\newcommand{\xmark}{\ding{55}}  

\begin{document}

\title{Turbo-GoDec: Exploiting the Cluster Sparsity Prior for Hyperspectral Anomaly Detection}

\author{Jiahui~Sheng, Xiaorun~Li and Shuhan~Chen
	
	\thanks{
	This manuscript is a preprint. The paper has been accepted for publication
	in IEEE Transactions on Multimedia. The final published version will be
	available on IEEE Xplore.
	}
}


\maketitle

\begin{abstract}
As a key task in hyperspectral image processing, hyperspectral anomaly detection has garnered significant attention and undergone extensive research. 
Existing methods primarily relt on two prior assumption: low-rank background and sparse anomaly, along with additional spatial assumptions of the background. 
However, most methods only utilize the sparsity prior assumption for anomalies and rarely expand on this hypothesis.
From observations of hyperspectral images, we find that anomalous pixels exhibit certain spatial distribution characteristics: they often manifest as small, clustered groups in space, which we refer to as cluster sparsity of anomalies.
Then, we combined the cluster sparsity prior with the classical GoDec algorithm, incorporating the cluster sparsity prior into the S-step of GoDec. This resulted in a new hyperspectral anomaly detection method, which we called Turbo-GoDec.
In this approach, we modeled the cluster sparsity prior of anomalies using a Markov random field and computed the marginal probabilities of anomalies through message passing on a factor graph. Locations with high anomalous probabilities were treated as the sparse component in the Turbo-GoDec.
Experiments are conducted on three real hyperspectral image (HSI) datasets which demonstrate the superior performance 
of the proposed Turbo-GoDec method in detecting small-size anomalies 
comparing with the vanilla GoDec (LSMAD) and state-of-the-art anomaly detection methods.
\dd{The code is available at https://github.com/jiahuisheng/Turbo-GoDec.}
\end{abstract}


\IEEEpeerreviewmaketitle

\section{Introduction}

\IEEEPARstart{U}{nlike} visible light images, hyperspectral images have hundreds of channels, which can provide spectral information about the objects \cite{background-1}, \cite{background-2}. 
Due to the unique data structure and rich spectral features of hyperspectral images, techniques such as hyperspectral target detection, classification, and unmixing have emerged \cite{usage-1, usage-2, usage-TD, CGCRD, SPSTT, TRGLF, usage-Classification}. 
Among these, hyperspectral anomaly detection is a prominent topic that enables the detection of anomalous targets without prior spectral information \cite{anomaly-1}, \cite{anomaly-2}.
In simple terms, the goal of anomaly detection can be seen as highlighting anomalies in the detection map while suppressing the background.

Since hyperspectral anomaly detection is an unsupervised task that does not require the spectral information of the targets, it is highly practical. However, the lack of prior spectral information for the targets also makes it challenging to clearly define anomalies versus non-anomalies, which has led to differing opinions among researchers regarding the definition of anomalies \cite{anomaly-def-1}, \cite{IGPAD}. Generally speaking, we consider anomalies to have two characteristics: significant spectral differences and occupying a very small number of pixels. Therefore, in this paper, we only focus on small-sized anomalies.

Over the past two decades, numerous researchers have devoted significant effort to this field, resulting in the development of many effective detection methods. Generally, these methods can be broadly categorized into four types: 
\begin{enumerate}[]
	
	\item Statistics-based methods, primarily represented by the Reed-Xiaoli anomaly detector (RXAD) \cite{GRX} and its variants, such as LRX \cite{LRX}, KRX \cite{KRX}, and FrRX \cite{FrRX}.
	
	\item Representation-based methods, which include techniques based on low-rank sparse representation \cite{LRASR,LRASR_BP,LRASR_UD,SD_LRASR,LTNN} and collaborative representation \cite{CRD, CR_PCA, RCRD,morCRD,swCRD}.

	\item Decomposition-based methods, further divided into matrix decomposition-based \cite{LSMAD,LRaSMD,MX-SVD,LSDM-MoG} and tensor decomposition-based approaches \cite{Tucker,TensorRPCA,FA_TD,PBTA}.

	\item Deep learning-based methods, encompassing techniques based on CNNs \cite{CNN-based}, GANs \cite{GAN-based}, Transformer \cite{Transformer-based}, and Autoencoders (AEs) \cite{Auto-AD, RGAE, Memory-Augmented-AE}.
	
\end{enumerate}

In addition to the four main categories of anomaly detection methods mentioned above, there are other methods proposed for hyperspectral anomaly detection, like topology-based methods \cite{topology-based}, \cite{chessboad-topology} and graph-based methods \cite{graph-based-1}, \cite{graph-based-2}.

Despite the use of various techniques and the proposal of numerous methods for hyperspectral anomaly detection, these approaches consistently rely on two prior assumptions: that the background is low-rank and that anomalies are sparse. Based on these two assumptions, researchers employ appropriate techniques to model or leverage these two priors. In this way, the overall approach of current methods can be roughly divided into two types: one is to seek newer, better, and more suitable models to model or leverage the low-rank assumption of background and sparse assumption of anomaly, which is the most common approach; the other is to consider more accurate and more realistic prior assumptions for hyperspectral images, which is relatively less common. 

In the category of methods that optimize prior assumptions, the vast majority focus on optimizing the prior assumptions about the background. In \cite{LRX} and \cite{CRD}, the authors argue that the background should be low-rank within a certain spatial range, which is a further extension of the low-rank assumption of background. Similarly, Gao et al. introduced blind spot networks \cite{BSN} into hyperspectral anomaly detection based on similar assumptions, achieving good detection results \cite{BS3LNet}, \cite{PDBSNet}. In \cite{TVSDM} and \cite{GTVLRR}, the total variation regularization is considered to guide the reconstruction of the background, which is based on the assumption that the background changes smoothly in space. 

In contrast, extensions of the sparse assumption of anomaly are considerably rarer. In \cite{ELRSF-SP}, Xiao et al. suggest that the transition between anomalies and the background should be abrupt, therefore introducing the saliency prior term to enhance the contrast between anomalies and the background, which focuses more on the spectral differences of anomalies. However, it is surprising that very few researchers consider the spatial characteristics of anomalies. If you observe the spatial distribution of anomalies, you can easily find that anomalous targets often occupy several pixels and cluster in small areas, which is referred to in this paper as the cluster sparsity prior assumption for anomalies. Once we employ the cluster sparsity prior for anomalies, we can compel our model to focus more on those small groups of anomalies. This can also enable the model to ignore large areas of false positive targets, significantly reducing false alarms. Additionally, the model will pay more attention to the surroundings of anomalous pixels, enhancing its capability to detect sub-pixel anomalies at the edges of anomalous targets, thereby reducing missed detections to some extent. So, why has no one utilized the cluster sparsity prior for anomalies? First, many researchers are overly focused on how to apply new technologies to existing assumptions, neglecting the optimization of the sparsity prior assumption for anomaly. Secondly, and more importantly, it is quite challenging to model the cluster sparsity prior for anomalies in hyperspectral images.

To effectively leverage the cluster sparsity prior for hyperspectral anomaly detection, we propose Turbo-GoDec, which is based on the GoDec matrix decomposition model. Our proposed Turbo-GoDec not only successfully models the cluster sparsity prior for anomalies but also provides a framework for other researchers looking to utilize the cluster prior hypothesis in their work. Additionally, as mentioned earlier, Turbo-GoDec can reduce large areas of false positives while enhancing the detection capability for sub-pixel anomalies at the edges of anomalous targets.

The workflow of Turbo-GoDec can be summarized as follows. First, we use the GoDec algorithm to perform matrix decomposition on the hyperspectral image. After obtaining the low-rank component from each iteration, we subtract this component from the original hyperspectral image, denoting the result as $T_t$. Next, we sum $T_t$ across the channel dimension to convert it into a scalar for each pixel. We then model the cluster sparsity prior using a Markov random field and connect this prior to $T_t$ through a factor graph. Once the factor graph model is established, we will perform message passing on the factor graph to compute the marginal probabilities of each pixel being anomalous. Based on the marginal probabilities, we derive the corresponding sparse components. This process is repeated until convergence. Finally, Turbo-GoDec outputs the low-rank component, sparse component, and anomalous probabilities of the hyperspectral image, allowing us to generate the corresponding detection map.

The main contributions of this paper can be summarized as follows:

\begin{enumerate}[]
	
	\item We innovatively considered the spatial distribution characteristics of anomalies, and to my knowledge, this is the first time that cluster sparsity prior has been utilized in hyperspectral anomaly detection.
	
	\item We successfully modeled the cluster sparsity prior using a Markov random field and represented the Markov random field with a factor graph. We then implemented a message passing algorithm on the factor graph to infer the marginal probabilities of individual pixels being anomalous.

	\item We combined GoDec with the cluster sparsity prior to propose Turbo-GoDec, which significantly enhances the detection of mixed pixels at anomaly edges and reduces false alarms. Moreover, our proposed Turbo-GoDec method also provides a paradigm for how to leverage cluster sparsity prior in hyperspectral anomaly detection.
	
\end{enumerate}

The rest of this paper is organized as follows. 
Section II provides a brief introduction to GoDec and factor graph. 
In Section III, we present a detailed description of our proposed Turbo-GoDec, including the overall framework of Turbo-GoDec, construction of the cluster sparsity model, and the inference on the factor graph. 
In Section IV, we conducted comparative experiments for Turbo-GoDec.
Finally, we concludes the paper in Section V.

\dd{
\section{Related Work}
Current hyperspectral anomaly detection methods can be roughly divided into four categories: 1) statistics-based method, 2) representation-based method, 3) decomposition-based method, and 4) deep learning-based method. Next, we will introduce related work based on these four categories.

\subsection{Statistics-based Method}
Statistical-based hyperspectral anomaly detection algorithms primarily detect anomalies by leveraging the statistical characteristics of hyperspectral data, including the classical Reed–Xiaoli (RX) algorithm \cite{GRX}, which is widely used as the benchmark in HAD field. The RX is based on the assumption that the background follow the multivariate Gaussian distribution, which deviates somewhat from real-world scenarios. 
Therefore, some variants of RXAD are proposed to alleviate this problem, such as the local RX \cite{LRX}, the weighted RX \cite{WRX} and the kernel RX \cite{KRX}. 

In addition to RX-based methods, several other statistical approaches have also been proposed, including the two-step GLRT (2S-GLRT) method \cite{2S-GLRT}, the distributed online one-class support vector machine (doCSVM) method \cite{doCSVM}, and the kernel isolation forest detection (KIFD) method \cite{KIFD}.

\subsection{Representation-based Method}
Due to the excellent interpretability and strong detection performance, representation-based methods have attracted significant attention from researchers. As the two main branches of representation-based methods, CRD \cite{CRD} and LRASR \cite{LRASR} have been extensively studied and have inspired numerous extended approaches, such as the weighted CRD (WCRD) \cite{swCRD}, morphology-based CRD \cite{morCRD} and the graph and total variation regularized LR representation (GTVLRR) \cite{GTVLRR}. To cope with the high computational cost of CRD, Ma et al. \cite{RCRD} propose a recursive CRD (RCRD) to reduce the computational complexity. In \cite{PCA-TLRSR}, Wang et al. first introduce the tensor low-rank representation into the HAD task, proposing a PCA-based Tensor Low-Rank and Sparse Representation (PCA-TLRSR) method.

\subsection{Decomposition-based Method}
Decomposition-based Method can be regarded as one special case of low-rank representation model, which intend to decompose the HSI into low-rank background component and sparse anomaly component. LSMAD \cite{LSMAD} and LRaSMD \cite{LRaSMD} are representative methods in this category, both of which are GoDec-based approaches. LSMAD is also the base model of our Turbo-GoDec. In addition to classical algorithms, several new decomposition-based methods have been proposed in recent years \cite{LSDM-MoG, TensorRPCA, PBTA}. In \cite{LSDM-MoG}, the LRaSMD is combined with Gaussian Mixture Model to better fit the data and reduce the mixing of anomalies and noise. Moreover, Ma et al. \cite{FrFT-RC-LRaSMD} propose a new LRaSMD-based method which integrate the fractional Fourier transform (FrFT) to separate the background and anomaly components.

\subsection{Deep learning-based Method}
Due to the powerful representation capabilities of deep learning models, they have been increasingly applied to hyperspectral anomaly detection (HAD) tasks. This includes methods based on autoencoders \cite{Auto-AD, RGAE, manifold-AE, TRPCA-AEAN, PSSIN}, GANs \cite{GAN-based,WeaklyAD}, transformers \cite{Transformer-based,GT-HAD, TAEF}, and diffusion models \cite{BSDM,DWSDiff}. 
As one of the pioneers of DNN-based HAD, AutoAD \cite{Auto-AD} employs a fully connected CNN-based autoencoder to reconstruct the background, which is commonly used as a benchmark for HAD. In \cite{GT-HAD}, a gated transformer network is proposed for HAD, which is used to exploit the spatial–spectral similarity. Moreover, in \cite{TAEF}, Wu et al. develop a transformer-based AE framework (TAEF) to accurately model the nonlinear mixing phenomenon in HSIs.

}

\section{Preliminaries}
In this section, preliminary knowledge of the proposed method will be introduced, including the GoDec algorithm and the factor graph.

\subsection{The Principle of GoDec Algorithm}
GoDec is an matrix decomposition algorithm originally developed for matrix completion \cite{GoDec}. 
The GoDec algorithm models the target object as the sum of a low-rank component, a sparse component, and noise, as shown in the following equation: 
\begin{equation}
	\label{GoDec_equation}
	X = L + S + N
\end{equation}
where $X$ is the object matrix, $L$ is the low-rank component, $S$ is the sparse component, and $N$ is the noise. 
Therefore, the problem addressed by GoDec can be considered an approximate "low-rank+sparse" decomposition problem, which can be described as follows: 
\begin{align}
	\label{GoDec_problem}
	\begin{aligned}
		\min_{L,S}\quad	&\|X-L-S\|_F^2\\
		s.t.\quad		&\text{rank}\left(L\right)\leq r,\\
						&\text{card}\left(S\right)\leq k.
	\end{aligned}
\end{align}

To further simplify the problem, the optimization problem shown in Eq. \ref{GoDec_problem} can be decomposed into the following two subproblems: 
\begin{equation}
	\begin{cases}
		&L_t=\arg\min_{\operatorname{rank}(L)\leq r}\|X-L-S_{t-1}\|_F^2;\\
		&S_t=\arg\min_{\operatorname{card}(S)\leq k}\|X-L_t-S\|_F^2.
	\end{cases}
\end{equation}
\dd{where $\operatorname{rank}(L)$ is the rank of matrix $L$, $\operatorname{card}(S)$ is the cardinality of matrix $S$ (the number of non-zero entry in $S$). $\|\cdot \|_F$ represents the Frobenius norm. $L_t$ and $S_t$ represent the low-rank and sparse components at the t-th iteration, respectively.}

The subproblem for the low-rank component can be solved using singular value decomposition, while the subproblem for the sparse component can be addressed with an entry-wise hard thresholding algorithm, 
\begin{equation}
	\label{eq: GoDec_solution}
	\begin{aligned}
		L_{t}&=\sum_{i=1}^{r}\lambda_{i}U_{i}V_{i}^{T},\mathrm{svd}\left(X-S_{t-1}\right)=U\Lambda V^{T};\\
		S_{t}&=\mathcal{P}_{\Omega}\left(X-L_{t}\right),\Omega:\left|\left(X-L_{t}\right)_{i,j\in\Omega}\right|\neq0\\
			 &\mathrm{and} \geq\left|\left(X-L_{t}\right)_{i,j\in\overline{\Omega}}\right|,\left|\Omega\right|\leq k.
	\end{aligned}
\end{equation}
\dd{where SVD is the singular value decomposition operation, $\lambda_{i}$ is the $i$'th largest singular value. $\mathcal{P}_{\Omega}$ represents the projection of a matrix onto an entry set $\Omega$, which means selecting the entries from $X-L_t$ according to the indices in $\Omega$. $\Omega$ represents the nonzero subset of the $k$ largest entries in $X-L_t$, and $\overline{\Omega}$ means the complementary set of $\Omega$.}

Considering that singular value decomposition can be time-consuming, $L_t$ can also be updated using bilinear random projection, which is called Fast GoDec \cite{GoDec}. 

Readers can refer to \cite{GoDec} for more information about GoDec algorithm.
Meanwhile, there is no need to delve too deeply into the update mechanism of $L_t$, since this paper mainly focuses on improvements in the update process of $S_t$.

\subsection{Factor Graph}
Factor graph is a type of probabilistic graphical model that can directly represent the factorization of probability distributions \cite{factor-graph}. For a multivariate joint probability distribution, a factor graph can explicitly represent its factorization. 

Formally, a factor graph consists of variable nodes and factor nodes connected by edges. In the graph, variable nodes represent random variables and are typically depicted as circles, while factor nodes represent probability distributions or functions and are usually represented as squares. 

Below is a simple example of a factor graph, considering the case of three random variables. Suppose their joint probability distribution satisfies the following equation: 
\begin{equation}
    \label{eq:factor_graph}
	p(x_1, x_2, x_3) = p(x_3|x_2, x_1) p(x_2|x_1) p(x_1)
\end{equation}
This joint distribution can be factored into three components: $f_1(x_1, x_2, x_3) = p(x_3|x_2, x_1)$, $f_2(x_1, x_2) = p(x_2|x_1)$, $f_3(x_1) = p(x_1)$. 
Therefore, this joint probability distribution can be modeled as the following factor graph:
\begin{figure}[!h]
\centering
\includegraphics[width=0.5\linewidth]{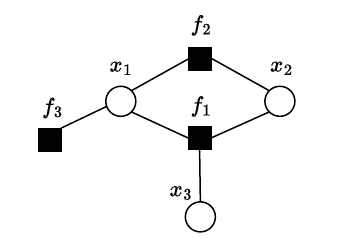}
\caption{
    An example of the factor graph. The joint probability is factored as the product of three factor nodes.
}
\label{fig:factor graph example}
\end{figure}

With the help of factor graphs, we can simplify the representation of complex models and streamline the inference process.

\section{Proposed Method}

In this section, we will first introduce the overall framework and process of Turbo-GoDec. Then, we will provide a detailed explanation of how we model cluster sparsity prior, followed by inference based on the message passing algorithm on the obtained factor graph. 

\subsection{Overall Framework of Turbo-GoDec}
\begin{figure*}[!ht]
\centering
\includegraphics[width=0.9\linewidth]{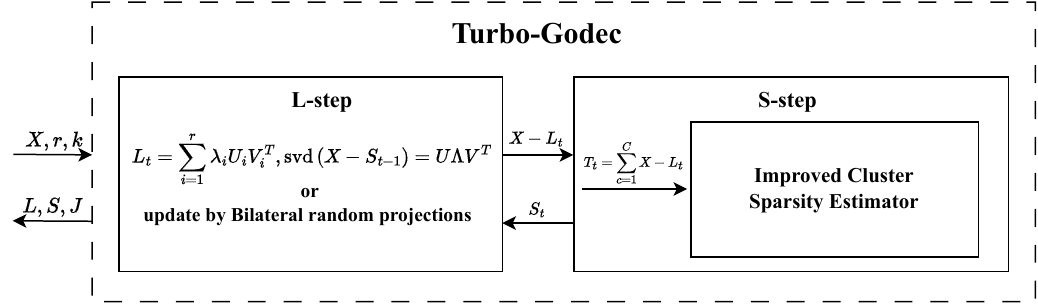}
\caption{
    The overall framework of Turbo-GoDec. As shown in the figure, Turbo-GoDec has three inputs: the hyperspectral image $X$, the rank constraint $r$ for the low-rank component, and the sparsity constraint (cardinality) $k$ for the sparse component. The process of Turbo-GoDec can be viewed as alternating updates of the L-step and S-step. The L-step computes the low-rank component, while the S-step calculates the sparse component. 
    Finally, after the iterations stop, Turbo-GoDec outputs the low-rank component $L$, the sparse component $S$, and the anomalous probability $J$ in the hyperspectral image.
}
\label{fig:framework}
\end{figure*}

In this section, we will introduce the overall framework of our proposed Turbo-GoDec. The framework is illustrated in Fig. \ref{fig:framework}. As shown in the Fig. \ref{fig:framework}, the overall process of Turbo-GoDec does not differ significantly from that of GoDec. The entire algorithm can be viewed as an alternating optimization of the L-step and S-step. By the way, we additionally consider the cluster sparsity prior for anomalies, which can be understood as updating our optimization problem to the following equation:
\begin{align}
	\label{Turbo_GoDec_problem}
	\begin{aligned}
		\min_{L,S}\quad	&\|X-L-S\|_F^2 + \text{CSP}(S)\\
		s.t.\quad		&\text{rank}\left(L\right)\leq r,\\
						&\text{card}\left(S\right)\leq k.
	\end{aligned}
\end{align}
$\text{CSP}$ represents the addition of a regularization term related to the cluster sparsity prior for the $S$ component. Moreover, the corresponding two subproblems are also modified to the following equations:
\begin{equation}
	\begin{cases}
		&L_t=\arg\min_{\operatorname{rank}(L)\leq r}\|X-L-S_{t-1}\|_F^2;\\
		&S_t=\arg\min_{\operatorname{card}(S)\leq k}\|X-L_t-S\|_F^2 + \text{CSP}(S).
	\end{cases}
\end{equation}

\subsubsection{L-step Update}
We can observe that, compared to the optimization problem \ref{GoDec_problem} in GoDec, the optimization problem \ref{Turbo_GoDec_problem} in Turbo-GoDec only adds a regularization term for the cluster sparsity prior on $S$. This means that the solutions for the low-rank components in both methods remain the same.
The low-rank component can be updated via the singular value hard thresholding of $X-S_{t-1}$ as follows:
\begin{equation}
	L_{t}=\sum_{i=1}^{r}\lambda_{i}U_{i}V_{i}^{T},\mathrm{svd}\left(X-S_{t-1}\right)=U\Lambda V^{T}
\end{equation}
where $S_0$ is set as a zero matrix.

As mentioned in the Preliminaries section, the low-rank component can also be updated using bilinear random projection, which helps reduce the computation cost.

\subsubsection{S-step Update}
Due to the additional consideration of the cluster sparsity prior for anomalies, the sparse component can no longer be updated using the hard thresholding algorithm from Eq. \ref{eq: GoDec_solution}. 

To address this, we modeled the cluster sparsity in the two-dimensional space using a Markov random field. Each iteration's $X-S_{t-1}$ serves as our observation, and we employed a message passing algorithm on the factor graph to compute the posterior probabilities of the anomalies. The specific implementation will be elaborated in the subsequent sections.

\begin{algorithm}[!t]
	\dd{
    \footnotesize
    \caption{The Pseudo-code of Turbo-GoDec}
    \begin{itemize}
    \item \textbf{Input:} Hyperpsectral Image $X$; potential function $\psi_{00}$, $\psi_{01}$, $\psi_{10}$, $\psi_{11}$; trade-off parameter $\alpha$; the error tolerance $\epsilon$; the GoDec iteration times $T_{G}$; the S-step iteration times $T_{S}$.
    \item \textbf{Output:} Low-rank Component $L$; Sparse Component $S$; Anomalous Probability $J$.
    \end{itemize}
    \begin{algorithmic}[1]

    \State Initiate: $L_{0} \leftarrow X$, $S_{0} \leftarrow 0$, $t \leftarrow 0$, $\gamma^{l} \leftarrow 0$, $\gamma^{r} \leftarrow 0$, $\gamma^{t} \leftarrow 0$, $\gamma^{b} \leftarrow 0$, $C_0^r = C_0^b = \psi_{00} + \psi_{10}$, $C_1^r = C_1^b =\psi_{11} + \psi_{01}$, $C_0^t = C_0^l = \psi_{00} + \psi_{01}$, $C_1^t = C_1^l = \psi_{11} + \psi_{10}$
	\State Obtain the rank $r$ and the cardinality $k$ through NWHFC and MX-SVD for $X$
    \While{$\|X-L_t-S_{t}\|_F^2 / \|X\| > \epsilon$ and $t<T_{G}$}
		\State $t \leftarrow t+1$
		\State $\lambda,U,V \leftarrow \mathrm{svd}\left(X-S_{t-1}\right)$
		\State $L_{t} \leftarrow \sum_{i=1}^{r}\lambda_{i}U_{i}V_{i}^{T}$
		\State $T_t \leftarrow \sum_{c=1}^{C} X-L_t$
		\State $\pi_{n,m}^{in} \leftarrow \left(1+\frac{\mathcal{N}\left(0;t_{n,m},\sigma_1^2\right)}{\mathcal{N}\left(0;t_{n,m},\sigma_1^2+\sigma_2^2\right)}\right)^{-1}$
        \For{$i = 1 $ to $T_{S}$}
			\State $\gamma_{n,m}^{l} \leftarrow \frac{ \psi_{01} (1-\pi_{n,m-1}^{in}) \alpha_{n,m-1}^{\cancel{r}} + \psi_{11}\pi_{n,m-1}^{in} \beta_{n,m-1}^{\cancel{r}}}
			{C_0^l (1-\pi_{n,m-1}^{in}) \alpha_{n,m-1}^{\cancel{r}}  +  C_1^l \pi_{n,m-1}^{in} \beta_{n,m-1}^{\cancel{r}}}$

            \State $\gamma_{n,m}^{r} \leftarrow \frac{ \psi_{10} (1-\pi_{n,m+1}^{in}) \alpha_{n,m}^{\cancel{l}} + \psi_{11}\pi_{n,m+1}^{in} \beta_{n,m}^{\cancel{l}}}
			{C_0^r (1-\pi_{n,m+1}^{in})\alpha_{n,m+1}^{\cancel{l}}  +  C_1^r \pi_{n,m+1}^{in} \beta_{n,m+1}^{\cancel{l}}}$

			\State $\gamma_{n,m}^{t} \leftarrow \frac{ \psi_{01} (1-\pi_{n-1,m}^{in}) \alpha_{n-1,m}^{\cancel{b}} + \psi_{11}\pi_{n-1,m}^{in} \beta_{n-1,m}^{\cancel{b}}}
			{C_0^t (1-\pi_{n-1,m}^{in})\alpha_{n-1,m}^{\cancel{b}}  +  C_1^t \pi_{n-1,m}^{in} \beta_{n-1,m}^{\cancel{b}}}$

			\State $\gamma_{n,m}^{b} \leftarrow \frac{ \psi_{10} (1-\pi_{n+1,m}^{in}) \alpha_{n+1,m}^{\cancel{t}} + \psi_{11}\pi_{n+1,m}^{in} \beta_{n+1,m}^{\cancel{t}}}
			{C_0^b (1-\pi_{n+1,m}^{in})\alpha_{n+1,m}^{\cancel{t}}  +  C_1^b \pi_{n+1,m}^{in} \beta_{n+1,m}^{\cancel{t}}}$

        \EndFor

		\State $\pi_{n,m}^{out} \leftarrow \frac
		{\prod_{d \in \{l,r,t,b\}} {\gamma_{n,m}^{d}}}
		{\prod_{d \in \{l,r,t,b\}} {1-\gamma_{n,m}^{d}} + \prod_{d \in \{l,r,t,b\}} {\gamma_{n,m}^{d}}}$

		\State $J_{n,m} \leftarrow p(a_{n,m} = 1) = \frac{\pi_{n,m}^{in} \pi_{n,m}^{out}}
		{\pi_{n,m}^{in} \pi_{n,m}^{out} + (1-\pi_{n,m}^{in}) (1-\pi_{n,m}^{out})}$
		
		\State $S_{t} \leftarrow \mathcal{P}_{\Omega}\left(X-L_{t}\right),\Omega:\left|J_{i,j\in\Omega}\right|\neq0 \ \mathrm{and} \geq\left|J_{i,j\in\overline{\Omega}}\right|,\left|\Omega\right|\leq k$
    \EndWhile 

    \end{algorithmic}
    \label{algm:Attraction and Repulsion Loss}
	}
\end{algorithm}

\begin{figure}[!ht]
\centering
\includegraphics[width=0.9\linewidth]{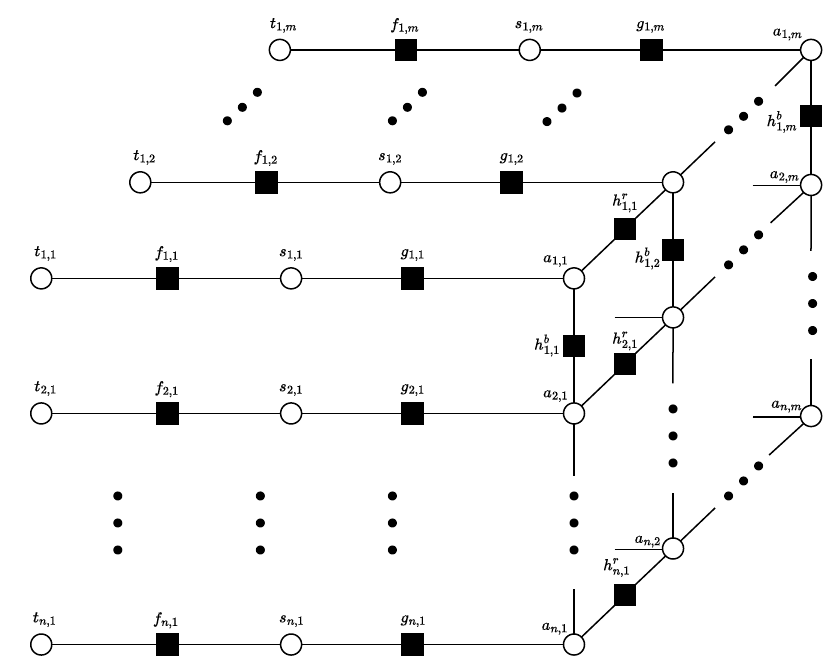}
\caption{
    The factor graph used in Turbo-GoDec. This factor graph is equivalent to the factorization of the joint probability expressed in Eq. \ref{eq:rewritte joint Posterior Probability}.}
\label{fig:factor graph}
\end{figure}

\subsection{Cluster Sparsity Prior Modelling}

To model the cluster sparsity prior for anomalies, we employed the factor graph representation of a Markov random field (MRF) \cite{MRF}, \cite{grid-MRF}. The concept is quite straightforward: we believe that the L-step of Turbo-GoDec effectively removes the background component from the hyperspectral image, leaving behind the sparse anomaly component and noise, which is typically considered Gaussian white noise. This way, the S-step can be viewed as detecting two-dimensional cluster sparsity signals in the presence of Gaussian white noise. 

To simplify the computation, we first sum the $X-L_t$ obtained from the L-step channel-wise, denoting this as $T_t \in R^{n \times m}$.
\begin{equation}
	T_t = \sum_{c=1}^{C} X-L_t
\end{equation}
Where $C$ here is the channel number of the hyperspectral image. As mentioned above, $T$ can be viewed as the sparse component plus Gaussian noise, so we can model it as follows: 
\begin{equation}
	\label{eq:T_model}
	T = S + Z,\ Z \sim \mathcal{N}(0, \sigma^2)
\end{equation}
Where $S$ represents the sparse component, which is mostly composed of zeros. 

To leverage the structural sparsity of anomalies, we introduced a support variable $A \in {0,1}^{n \times m}$ in the model, which indicates whether an anomaly is present at the corresponding position. Specifically, $a=1$ indicates the presence of an anomaly, while $a=0$ indicates no anomaly. 
Then, the joint probability of $T,S$ and $A$ is expressed as: 
\begin{equation}
	\label{eq:joint Posterior Probability}
	p(T,S,A)= p(T|S)p(S|A)p(A)
\end{equation}

Based on Eq. \ref{eq:T_model}, we can find $T$ is only related to the corresponding position of $S$, so the conditional probability distribution $p(T|S)$ can be expressed as:
\begin{equation}
	\label{eq:TS_cpd}
	p(T|S)= \prod_{n=1}^{N} \prod_{m=1}^{M} p(t_{n,m}|S)
	=\prod_{n=1}^{N} \prod_{m=1}^{M} p(t_{n,m}|s_{n,m})
\end{equation}
where the $p(t_{n,m}|s_{n,m})$ is the conditional probability distribution of each element, which is given by:
\begin{equation}
	\label{eq:ts_cpd}
	p(t_{n,m}|s_{n,m}) =\mathcal{N}(t_{n,m}; s_{n,m}, \sigma_1^2)
\end{equation}
which is denoted as $f_{n,m}$ in Fig. \ref{fig:factor graph}.

Similarly, the conditional probability distribution of $S$ can be expressed as:
\begin{equation}
	\label{eq:SA_cpd}
	p(S|A)= \prod_{n=1}^{N} \prod_{m=1}^{M} p(S_{n,m}|A)
	=\prod_{n=1}^{N} \prod_{m=1}^{M} p(S_{n,m}|A_{n,m})
\end{equation}
As a support variable, $A$ significantly influences the distribution of the variable 
$S$ in the corresponding position, which then is given by the following equation \cite{Sparsity_model}:
\begin{equation}
	\label{eq:sa_cpd}
	p(s|a) = (1-a)\delta(s) + a\mathcal{N}(s; 0, \sigma_2^2)
\end{equation}
where $\delta$ is the Dirac delta function. Eq. \ref{eq:sa_cpd} indicates that when the support variable $a$ is 0, the value of $s$ is 0, and when $a$ is 1, the value of $s$ follows a Gaussian distribution. By the way, Eq. \ref{eq:sa_cpd} is denoted as $g$ in Fig. \ref{fig:factor graph}.

\begin{figure}[!h]
    \centering
    
    \begin{subfigure}{0.4\linewidth}
        \centering
        \includegraphics[width=\linewidth]{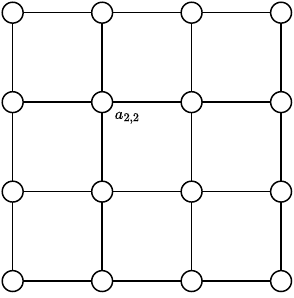}
        \caption{The 4$\times$4 grid-like Markov Random Field.}
        \label{fig:MRF}
    \end{subfigure}
    \hspace{0.05\linewidth}
    \begin{subfigure}{0.4\linewidth}
        \centering
        \includegraphics[width=\linewidth]{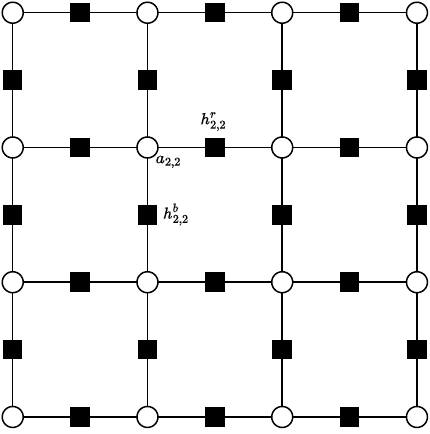}
        \caption{The corresponding factor graph.}
        \label{fig:MRF_factor_graph}
    \end{subfigure}
    
    \caption{Markov Random Field and its factor graph.}
    \label{fig:MRF_with_factor_graph}
\end{figure}

Regarding the support variable $A$, we aim to capture the cluster sparsity prior of anomalies in the two-dimensional plane. To achieve this, we employed a grid-like Markov random field for modeling, as illustrated in Fig. \ref{fig:MRF_with_factor_graph} along with its corresponding factor graph.

For the Markov random field shown in Fig. \ref{fig:MRF}, all adjacent pairs of points form the maximal cliques in the graph. According to the Hammersley-Clifford theorem, the joint probability of the $N \times M$ support variables can be factored as follows:
\begin{equation}
	\label{eq:A_pd}
	p(A) = p(a_1,a_2,\dots,a_{n\times M})=\frac{1}{Z}\prod_{c \in \mathcal{C}}\psi_c (a_c)
\end{equation}
where $Z$ is the partition function, $\mathcal{C}$ is the set of maximal cliques in the Markov random field, and $\psi_c (a_c)$ is the potential function defined on the maximal cliques, which is denoted as $h$ in the factor graph of Fig. \ref{fig:MRF_factor_graph}. 

Combining Eq. \ref{eq:TS_cpd}, \ref{eq:ts_cpd}, \ref{eq:SA_cpd}, \ref{eq:sa_cpd}, and \ref{eq:A_pd}, we can rewrite Eq. \ref{eq:joint Posterior Probability} as:
\begin{align}
	\label{eq:rewritte joint Posterior Probability}
	\begin{aligned}
	p(T,S,A) &= \frac{1}{Z}\prod_{c \in \mathcal{C}}\psi_c (a_c) \prod_{n=1}^{N} \prod_{m=1}^{M} \mathcal{N}(t_{n,m}; s_{n,m}, \sigma_1^2) \\
	& [(1-a_{n,m})\delta(s_{n,m}) + a_{n,m} \mathcal{N}(s_{n,m}; 0, \sigma_2^2)] 
	\end{aligned}
\end{align}

Based on the factorization of the joint distribution described in Eq. \ref{eq:rewritte joint Posterior Probability}, we can obtain the corresponding factor graph as shown in Fig. \ref{fig:factor graph}. 

\subsection{Message Passing on the Factor Graph}
After obtaining the factor graph, we aim to use it to compute the marginal probabilities of each support variable ($a$). To achieve this, we introduced a message passing algorithm that enables the parallel computation of the marginal probabilities for multiple support variables on the factor graph \cite{factor-graph}. Readers can refer to \cite{Turbo-CS}, \cite{Sparsity_model_2}, \cite{Sparsity_model_2}, and \cite{grid-MRF} for more information on the use of message passing. You can skip the specific calculations of message passing.

In this factor graph, the message passing can be divided into three main parts: $1)\ f_{n,m} \rightarrow s_{n,m} \rightarrow g_{n,m} \rightarrow a_{n,m}$, $2)$ Markov random field of $a$, and $3)\ a_{n,m} \rightarrow g_{n,m} \rightarrow s_{n,m} \rightarrow f_{n,m}$. The specific derivation results are as follows: 

1) Message Passing on the Path $f_{n,m} \rightarrow s_{n,m} \rightarrow g_{n,m} \rightarrow a_{n,m}$:

The message from factor node $f_{n,m}$ to variable node $s_{n,m}$ is 
\begin{equation}
    \label{eq:message_passing_on_f_s}
	m_{f_{n,m} \rightarrow s_{n,m}} = \mathcal{N}(t_{n,m}; s_{n,m}, \sigma_1^2)
\end{equation}

The message from variable node $s_{n,m}$ to factor node $g_{n,m}$ is the same as the message from factor node $f_{n,m}$ to variable node $s_{n,m}$.
\begin{equation}
    \label{eq:message_passing_on_s_g}
	m_{s_{n,m} \rightarrow g_{n,m}} = m_{f_{n,m} \rightarrow s_{n,m}} = \mathcal{N}(t_{n,m}; s_{n,m}, \sigma_1^2)
\end{equation}

The message from factor node $g_{n,m}$ to variable node $a_{n,m}$ is 
\begin{align}
    \label{eq:message_passing_on_g_a}
	\begin{aligned}
	m_{g_{n,m} \rightarrow a_{n,m}} 
	& = \int {m_{s_{n,m} \rightarrow g_{n,m}} \times g_{n,m}(s_{n,m},a_{n,m})}  \,ds_{n,m} \\
	& = \pi_{n,m}^{in} \delta(a_{n,m}-1) + (1-\pi_{n,m}^{in}) \delta(a_{n,m})
	\end{aligned}
\end{align}
where 
\begin{align}
    \label{eq:pi_in}
	\begin{aligned}
	\pi_{n,m}^{in}=\left(1+\frac{\mathcal{N}\left(0;t_{n,m},\sigma_1^2\right)}{\mathcal{N}\left(0;t_{n,m},\sigma_1^2+\sigma_2^2\right)}\right)^{-1}
	\end{aligned}
\end{align}

2) Message Passing on the Markov random field of $A$:

The message from factor node $h_{n,m-1}^{r}$ to variable node $a_{n,m}$ is 
\begin{align}
    \label{eq:message_passing_on_MRF_left}
	\begin{aligned}
	m_{h_{n,m-1}^{r} \rightarrow a_{n,m}} 
	&= \sum_{a_{n,m-1}} m_{g_{n,m-1} \rightarrow a_{n,m-1}} 
	m_{h_{n,m-1}^{b} \rightarrow a_{n,m-1}} \\
	&m_{h_{n-1,m-1}^{b} \rightarrow a_{n,m-1}}
	m_{h_{n,m-2}^{r} \rightarrow a_{n,m-1}}
	h_{n,m-1}^{r}\\
	&= \gamma_{n,m}^{l} a_{n,m} + (1-\gamma_{n,m}^{l})(1-a_{n,m})
	\end{aligned}
\end{align}
where
\begin{align}
    \label{eq:gamma_left}
	\begin{aligned}
	\gamma_{n,m}^{l}=\frac{ \psi_{01} (1-\pi_{n,m-1}^{in}) \alpha_{n,m-1}^{\cancel{r}} + \psi_{11}\pi_{n,m-1}^{in} \beta_{n,m-1}^{\cancel{r}}}
	{C_0^l (1-\pi_{n,m-1}^{in}) \alpha_{n,m-1}^{\cancel{r}}  +  C_1^l \pi_{n,m-1}^{in} \beta_{n,m-1}^{\cancel{r}}}
	\end{aligned}
\end{align}
with
\begin{align}
    \label{eq:C_left_0}
	\begin{aligned}
	C_0^l = \psi_{00} + \psi_{01}
	\end{aligned}
\end{align}
\begin{align}
    \label{eq:C_left_1}
	\begin{aligned}
	C_1^l = \psi_{11} + \psi_{10}
	\end{aligned}
\end{align}
\begin{align}
    \label{eq:alpha_left}
	\begin{aligned}
	\alpha_{n,m-1}^{\cancel{r}}=\prod_{d \in{l,t,b}} {1-\gamma_{n,m-1}^{d}}
	\end{aligned}
\end{align}
\begin{align}
    \label{eq:beta_left}
	\begin{aligned}
	\beta_{n,m-1}^{\cancel{r}}=\prod_{d \in{l,t,b}} {\gamma_{n,m-1}^{d}}
	\end{aligned}
\end{align}
The $\psi$ here is the potential function defined on the adjacent pixels.

Similarly, the message from other three factor node to variable node $a_{n,m}$ is 
\begin{align}
    \label{eq:message_passing_on_MRF_right}
	\begin{aligned}
	m_{h_{n,m}^{r} \rightarrow a_{n,m}} 
	= \gamma_{n,m}^{r} a_{n,m} + (1-\gamma_{n,m}^{r})(1-a_{n,m})
	\end{aligned}
\end{align}
\begin{align}
    \label{eq:message_passing_on_MRF_top}
	\begin{aligned}
	m_{h_{n-1,m}^{b} \rightarrow a_{n,m}} 
	= \gamma_{n,m}^{t} a_{n,m} + (1-\gamma_{n,m}^{t})(1-a_{n,m})
	\end{aligned}
\end{align}
\begin{align}
    \label{eq:message_passing_on_MRF_bottom}
	\begin{aligned}
	m_{h_{n,m}^{b} \rightarrow a_{n,m}} 
	= \gamma_{n,m}^{b} a_{n,m} + (1-\gamma_{n,m}^{b})(1-a_{n,m})
	\end{aligned}
\end{align}
where
\begin{align}
    \label{eq:gamma_right}
	\begin{aligned}
	\gamma_{n,m}^{r}=\frac{ \psi_{10} (1-\pi_{n,m+1}^{in}) \alpha_{n,m}^{\cancel{l}} + \psi_{11}\pi_{n,m+1}^{in} \beta_{n,m}^{\cancel{l}}}
	{C_0^r (1-\pi_{n,m+1}^{in})\alpha_{n,m+1}^{\cancel{l}}  +  C_1^r \pi_{n,m+1}^{in} \beta_{n,m+1}^{\cancel{l}}}
	\end{aligned}
\end{align}
\begin{align}
    \label{eq:gamma_top}
	\begin{aligned}
	\gamma_{n,m}^{t}=\frac{ \psi_{01} (1-\pi_{n-1,m}^{in}) \alpha_{n-1,m}^{\cancel{b}} + \psi_{11}\pi_{n-1,m}^{in} \beta_{n-1,m}^{\cancel{b}}}
	{C_0^t (1-\pi_{n-1,m}^{in})\alpha_{n-1,m}^{\cancel{b}}  +  C_1^t \pi_{n-1,m}^{in} \beta_{n-1,m}^{\cancel{b}}}
	\end{aligned}
\end{align}
\begin{align}
    \label{eq:gamma_bottom}
	\begin{aligned}
	\gamma_{n,m}^{b}=\frac{ \psi_{10} (1-\pi_{n+1,m}^{in}) \alpha_{n+1,m}^{\cancel{t}} + \psi_{11}\pi_{n+1,m}^{in} \beta_{n+1,m}^{\cancel{t}}}
	{C_0^b (1-\pi_{n+1,m}^{in})\alpha_{n+1,m}^{\cancel{t}}  +  C_1^b \pi_{n+1,m}^{in} \beta_{n+1,m}^{\cancel{t}}}
	\end{aligned}
\end{align}
with
\begin{align}
    \label{eq:C_right_0}
	\begin{aligned}
	C_0^r = C_0^b = \psi_{00} + \psi_{10}
	\end{aligned}
\end{align}
\begin{align}
    \label{eq:C_right_1}
	\begin{aligned}
	C_1^r = C_1^b =\psi_{11} + \psi_{01}
	\end{aligned}
\end{align}
\begin{align}
    \label{eq:C_top_0}
	\begin{aligned}
	C_0^t = C_0^l = \psi_{00} + \psi_{01}
	\end{aligned}
\end{align}
\begin{align}
    \label{eq:C_top_1}
	\begin{aligned}
	C_1^t = C_1^l = \psi_{11} + \psi_{10}
	\end{aligned}
\end{align}
and 
\begin{align}
    \label{eq:alpha_general}
	\begin{aligned}
	\alpha_{n,m}^{\cancel{o}}=\prod_{d \in \{l,r,t,b\} \setminus o} {1-\gamma_{n,m}^{d}}
	\end{aligned}
\end{align}
\begin{align}
    \label{eq:beta_general}
	\begin{aligned}
	\beta_{n,m}^{\cancel{o}}=\prod_{d \in \{l,r,t,b\} \setminus o} {\gamma_{n,m}^{d}}
	\end{aligned}
\end{align}

Due to the presence of numerous cycles in the Markov random field, message passing on the Markov random field requires iterations to converge to the correct values.

3) Message Passing on the Path $a_{n,m} \rightarrow g_{n,m} \rightarrow s_{n,m} \rightarrow f_{n,m}$:

The message from variable node $a_{n,m}$ to factor node $g_{n,m}$ is 
\begin{align}
    \label{eq:message_passing_on_a_g}
	\begin{aligned}
	m_{a_{n,m} \rightarrow g_{n,m}}
	&= m_{h_{n,m-1}^{r} \rightarrow a_{n,m}}
	 m_{h_{n,m}^{r} \rightarrow a_{n,m}} \\
	&\times m_{h_{n-1,m}^{b} \rightarrow a_{n,m}}
	 m_{h_{n,m}^{b} \rightarrow a_{n,m}} \\
	& = \pi_{n,m}^{out} \delta(a_{n,m}-1) + (1-\pi_{n,m}^{out}) \delta(a_{n,m})
	\end{aligned}
\end{align}
where 
\begin{align}
    \label{eq:pi_out}
	\begin{aligned}
	\pi_{n,m}^{out}=\frac
	{\prod_{d \in \{l,r,t,b\}} {\gamma_{n,m}^{d}}}
	{\prod_{d \in \{l,r,t,b\}} {1-\gamma_{n,m}^{d}} + \prod_{d \in \{l,r,t,b\}} {\gamma_{n,m}^{d}}}
	\end{aligned}
\end{align}

The message from factor node $g_{n,m}$ to variable node $s_{n,m}$ is 
\begin{align}
    \label{eq:message_passing_on_g_s}
	\begin{aligned}
	m_{g_{n,m} \rightarrow s_{n,m}}
	= \sum_{a_{n,m}} m_{a_{n,m} \rightarrow g_{n,m}} g_{n,m}(s_{n,m},a_{n,m})
	\end{aligned}
\end{align}

4) Calculate Marginal Probability of the Support Varibales $a$:

After message passing, we can obtain the marginal probabilities of the $N \times M$ support variables using the following equation: 
\begin{align}
    \label{eq:calculate_marginal_probability}
	\begin{aligned}
	p(a_{n,m}) \propto m_{g_{n,m} \rightarrow a_{n,m}} \times 
	m_{a_{n,m} \rightarrow g_{n,m}}
	\end{aligned}
\end{align}
thus
\begin{align}
    \label{eq:calculate_marginal_probability_1}
	\begin{aligned}
	p(a_{n,m} = 1) = \frac{\pi_{n,m}^{in} \pi_{n,m}^{out}}
	{\pi_{n,m}^{in} \pi_{n,m}^{out} + (1-\pi_{n,m}^{in}) (1-\pi_{n,m}^{out})}
	\end{aligned}
\end{align}
which represents the probability of that pixel being an anomaly after considering the cluster sparsity prior.
 
\subsection{Obtain Anomaly Detection Map}
Using Eq. \ref{eq:calculate_marginal_probability_1}, we can obtain a matrix that records the probabilities of each pixel being an anomaly, denoted as $J$. 
In this way, we can apply the entry-wise hard thresholding algorithm to $J$ to obtain the indices corresponding to the sparse component, as expressed in the following equation: 
\begin{equation}
	\label{eq: Turbo_GoDec_S_solution}
	\begin{aligned}
		S_{t}&=\mathcal{P}_{\Omega}\left(X-L_{t}\right),\Omega:\left|J_{i,j\in\Omega}\right|\neq0\\
			 &\mathrm{and} \geq\left|J_{i,j\in\overline{\Omega}}\right|,\left|\Omega\right|\leq k.
	\end{aligned}
\end{equation}

We then pass $S_t$ to the L-step, performing iterative loops until convergence or until the maximum number of iterations is reached. Ultimately, as shown in Fig. \ref{fig:framework}, we can obtain the low-rank component $L$, the sparse component $S$, and the probability $J$ of each pixel being an anomaly in the hyperspectral image.

Similar to LSMAD \cite{LSMAD}, we treat the low-rank component as a cleaner background and use the RX algorithm \cite{GRX} for anomaly detection. Additionally, since our $J$ also represents the degree of anomaly for each pixel to some extent, we sum both to obtain our final anomaly detection results, as shown in the following equation:
\begin{equation}
	\label{eq:anomaly_map}
	\begin{aligned}
		AD_{map}=\alpha \text{RX}(X, L) + (1-\alpha) J
	\end{aligned}
\end{equation}
where $\alpha$ is a trade-off parameter.

\section{Experiments and Analysis}
In this section, we will evaluate the detection performance of Turbo-GoDec on real datasets.

\subsection{Experimental Setup}

\begin{figure}[!h]
\includegraphics[width=0.95\linewidth]{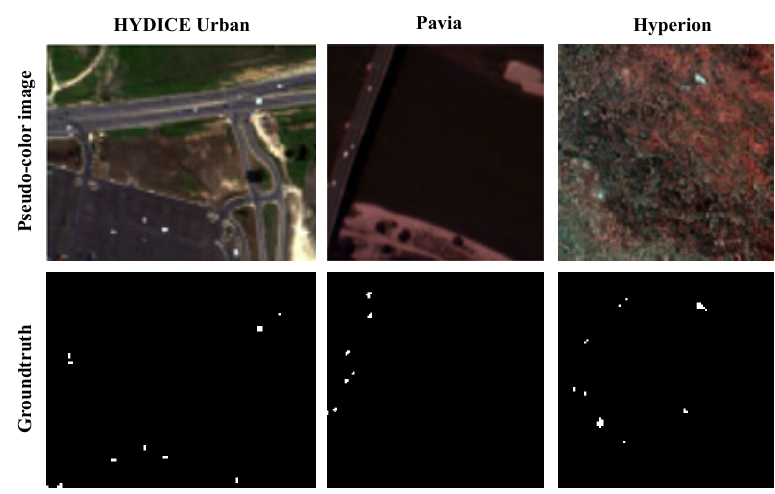}
\caption{
    The pseudo-color image and Groundtruth of three hyperspectral datasets.
}
\label{fig:dataset}
\end{figure}

\begin{figure*}[!ht]
\centering
\includegraphics[width=1\linewidth]{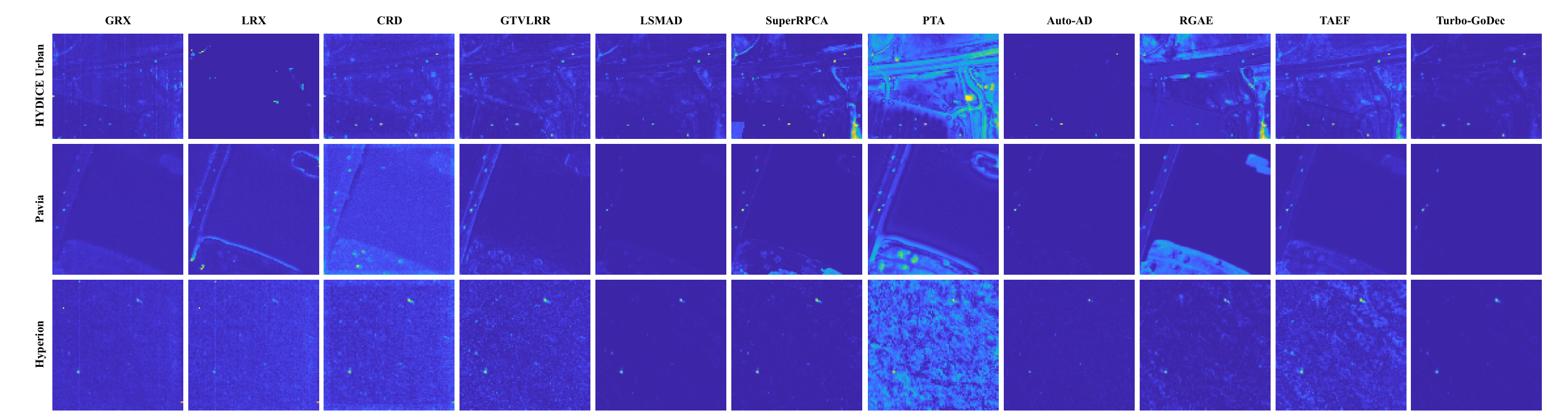}
\caption{
    Hyperspectral anomaly detection maps of comparison methods and proposed Turbo-GoDec method on three real hyperspectral datasets.
}
\label{fig:detection map}
\end{figure*}

\begin{figure*}[!ht]
\centering
\includegraphics[width=0.97\linewidth]{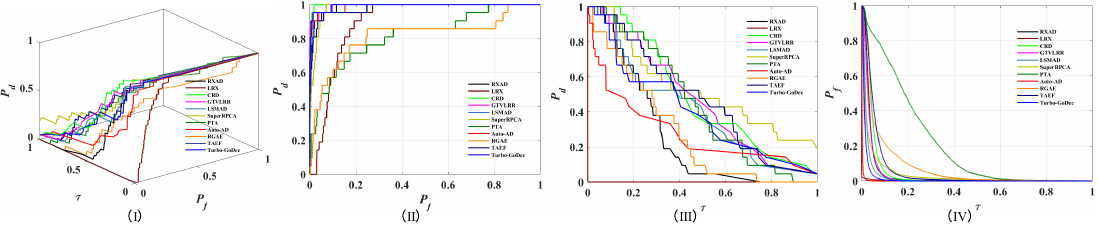}
\caption{The ROC curves of eleven anomaly detection methods on HYDICE Urban datasets. 
(\uppercase\expandafter{\romannumeral1}) 3D-ROC curve. 
(\uppercase\expandafter{\romannumeral2}) ROC curve of ($P_d-P_f$). 
(\uppercase\expandafter{\romannumeral3}) ROC curve of ($P_d-\tau$). 
(\uppercase\expandafter{\romannumeral4}) ROC curve of ($P_f-\tau$).}
\label{fig:HU_roc}
\end{figure*}

\begin{figure*}[!ht]
\centering
\includegraphics[width=0.97\linewidth]{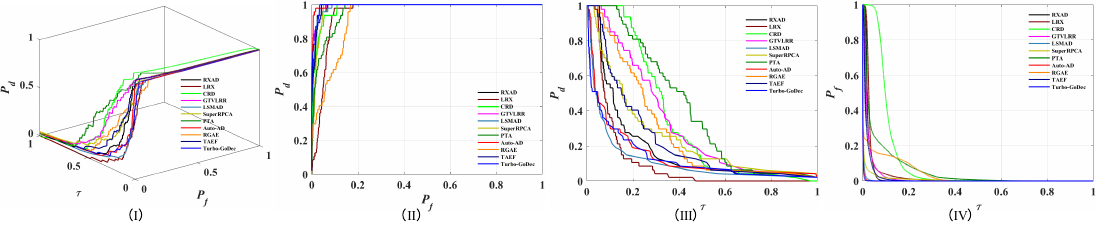}
\caption{The ROC curves of eleven anomaly detection methods on Pavia datasets. 
(\uppercase\expandafter{\romannumeral1}) 3D-ROC curve. 
(\uppercase\expandafter{\romannumeral2}) ROC curve of ($P_d-P_f$). 
(\uppercase\expandafter{\romannumeral3}) ROC curve of ($P_d-\tau$). 
(\uppercase\expandafter{\romannumeral4}) ROC curve of ($P_f-\tau$).}
\label{fig:Pavia_roc}
\end{figure*}

\begin{figure*}[!ht]
\centering
\includegraphics[width=0.97\linewidth]{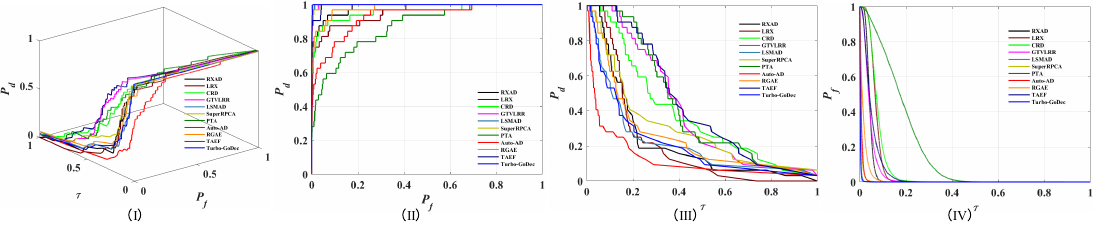}
\caption{The ROC curves of eleven anomaly detection methods on Hyperion datasets. 
(\uppercase\expandafter{\romannumeral1}) 3D-ROC curve. 
(\uppercase\expandafter{\romannumeral2}) ROC curve of ($P_d-P_f$). 
(\uppercase\expandafter{\romannumeral3}) ROC curve of ($P_d-\tau$). 
(\uppercase\expandafter{\romannumeral4}) ROC curve of ($P_f-\tau$).}
\label{fig:Hyperion_roc}
\end{figure*}

\begin{table*}[!ht]
	\renewcommand{\arraystretch}{1.2}
	\centering
	\caption{The values of AUC metrics of eleven methods on HYDICE Urban dataset.}
	\label{tab:HU_metric}
	\begin{tabular}{p{20mm}<{\centering}p{15mm}<{\centering}p{15mm}<{\centering}p{15mm}<{\centering}p{15mm}<{\centering}p{15mm}<{\centering}p{15mm}<{\centering}p{15mm}<{\centering}p{15mm}<{\centering}}
	\toprule
	$\text{Method}$ 	& $\text{AUC}_{\text{(D,F)}}$ & $\text{AUC}_{(\text{D},\tau)}$ & $\text{AUC}_{(\text{F},\tau)}$ & $\text{AUC}_{\text{TD}}$ & $\text{AUC}_{\text{BS}}$ & $\text{AUC}_{\text{SNPR}}$ & $\text{AUC}_{\text{TD-BS}} $& $\text{AUC}_{\text{ODP}}$ \\\midrule
	GRX 	    &0.9872				&0.2641				&0.0361				&1.2514				&0.9511				&7.3144				&0.2280				&1.2153	\\ 
	\bb{LRX} 	    &\bb{0.8947}			&\bb{0.0002}				&\bb{0.0024}			&\bb{0.8949}				&\bb{0.8923}				&\bb{0.0775}				&\bb{-0.0022}				&\bb{0.8924}			\\
	CRD 	    &\textbf{0.9978}	&\textbf{0.4739}	&0.0472				&\textbf{1.4716}	&0.9506				&10.0393			&\textbf{0.4267}	&\textbf{1.4244}\\
	GTVLRR 	    &0.9922				&\underline{0.4541}	&0.0331				&\underline{1.4463}	&0.9591				&13.7004			&\underline{0.4209}	&\underline{1.4131}\\
	LSMAD 	    &0.9925				&0.3920				&0.0221				&1.3845				&0.9704				&17.7096			&0.3699				&1.3624			\\
	\bb{SuperRPCA} 	    &\bb{0.9779}			&\bb{0.4239}				&\bb{0.0355}			&\bb{1.4018}				&\bb{0.9424}				&\bb{11.9480}				&\bb{0.3884}				&\bb{1.3664}			\\
	PTA 	    &0.8175				&0.4478				&0.2081				&1.2653				&0.6095				&2.1518				&0.2397				&1.0572			\\
	Auto-AD     &0.9899				&0.2661				&\textbf{0.0025}	&1.2560				&\textbf{0.9874}	&\textbf{107.8364}	&0.2637				&1.2536			\\
	RGAE 	    &0.8155				&0.2741				&0.0781				&1.0895				&0.7373				&3.5092				&0.1960				&1.0114			\\
	\bb{TAEF} 	    &\bb{0.9783}			&\bb{0.4522}				&\bb{0.0574}			&\bb{1.4305}				&\bb{0.9209}				&\bb{7.8844}				&\bb{0.3949}				&\bb{1.3732}			\\
	Turbo-GoDec	&\underline{0.9934}	&0.3875	            &\underline{0.0145}	&1.3809	            &\underline{0.9788}	&\underline{26.6635}&0.3730	            &1.3663         \\
	\bottomrule
	\end{tabular}
\end{table*}

\begin{table*}[!ht]
	\renewcommand{\arraystretch}{1.2}
	\centering
	\caption{The values of AUC metrics of eleven methods on Pavia dataset.}
	\label{tab:Pavia_metric}
	\begin{tabular}{p{20mm}<{\centering}p{15mm}<{\centering}p{15mm}<{\centering}p{15mm}<{\centering}p{15mm}<{\centering}p{15mm}<{\centering}p{15mm}<{\centering}p{15mm}<{\centering}p{15mm}<{\centering}}
	\toprule
	$\text{Method}$ 	& $\text{AUC}_{\text{(D,F)}}$ & $\text{AUC}_{(\text{D},\tau)}$ & $\text{AUC}_{(\text{F},\tau)}$ & $\text{AUC}_{\text{TD}}$ & $\text{AUC}_{\text{BS}}$ & $\text{AUC}_{\text{SNPR}}$ & $\text{AUC}_{\text{TD-BS}} $& $\text{AUC}_{\text{ODP}}$ \\\midrule
	GRX		        &0.9905				&0.1730				&0.0233				&1.1635				&0.9672				&7.4149					&0.1497				&1.1402\\
	\bb{LRX} 	    &\bb{0.9496}			&\bb{0.1109}				&\bb{0.0342}			&\bb{1.0604}				&\bb{0.9154}				&\bb{3.2431}				&\bb{0.0767}				&\bb{1.0263}\\
	CRD		        &0.9811				&\underline{0.3487}	&0.1092				&\underline{1.3298}	&0.8719				&3.1922					&0.2395				&1.2206\\
	GTVLRR	        &\underline{0.9935}	&0.3173				&0.0222				&1.3109				&0.9714				&14.3233				&\underline{0.2952}	&\underline{1.2887}\\
	LSMAD	        &0.9893				&0.1044				&0.0021	            &1.0936			    &0.9871				&49.1270				&0.1022				&1.0915\\
	\bb{SuperRPCA} 	    &\bb{0.9820}			&\bb{0.2210}				&\bb{0.0089}			&\bb{1.2029}				&\bb{0.9731}				&\bb{24.9418}				&\bb{0.2121}				&\bb{1.1941}			\\
	PTA		        &0.9635				&\textbf{0.3952}	&0.0590				&\textbf{1.3588}	&0.9045				&6.7001					&\textbf{0.3362}	&\textbf{1.2998}\\
	Auto-AD	        &\textbf{0.9966}	&0.1364				&\underline{0.0013}	&1.1329				&\textbf{0.9953}	&\underline{104.3319}		&0.1351				&1.1316\\
	RGAE	        &0.9277				&0.2678				&0.0380				&1.1955				&0.8896				&7.0451					&0.2298				&1.1575\\
	\bb{TAEF} 	    &\bb{0.9900}			&\bb{0.2214}				&\bb{0.0172}			&\bb{1.2114}				&\bb{0.9728}				&\bb{12.8838}				&\bb{0.2042}				&\bb{1.1942}\\
	Turbo-GoDec		&\underline{0.9935}	            &0.1231	            &\textbf{0.0011}	&1.1166	            &\underline{0.9923}	&\textbf{109.6896}	&0.1220	            &1.1155\\
	\bottomrule
	\end{tabular}
\end{table*}

\begin{table*}[!ht]
	\renewcommand{\arraystretch}{1.2}
	\centering
	\caption{The values of AUC metrics of eleven methods on Hyperion dataset.}
	\label{tab:Hyperion_metric}
	\begin{tabular}{p{20mm}<{\centering}p{15mm}<{\centering}p{15mm}<{\centering}p{15mm}<{\centering}p{15mm}<{\centering}p{15mm}<{\centering}p{15mm}<{\centering}p{15mm}<{\centering}p{15mm}<{\centering}}
	\toprule
	$\text{Method}$ 	& $\text{AUC}_{\text{(D,F)}}$ & $\text{AUC}_{(\text{D},\tau)}$ & $\text{AUC}_{(\text{F},\tau)}$ & $\text{AUC}_{\text{TD}}$ & $\text{AUC}_{\text{BS}}$ & $\text{AUC}_{\text{SNPR}}$ & $\text{AUC}_{\text{TD-BS}} $& $\text{AUC}_{\text{ODP}}$ \\\midrule
	GRX		        &0.9829				&0.2319				&0.0435				&1.2148				&0.9395				&5.3358				&0.1884				&1.1714\\
	\bb{LRX} 	    &\bb{0.9525}			&\bb{0.2136}				&\bb{0.0693}			&\bb{1.1661}				&\bb{0.8831}				&\bb{3.0814}				&\bb{0.1443}				&\bb{1.0968} \\
	CRD		        &0.9574				&0.3678				&0.0768				&1.3252	&0.8806				&4.7866				&0.2910		 &1.2484\\
	GTVLRR	        &0.9980	            &0.3969	&0.0440				&\underline{1.3949}	&0.9540				&9.0222				&\underline{0.3529}	&\underline{1.3509}\\
	LSMAD	        &\underline{0.9992}	&0.2059				&\underline{0.0030}	&1.2051				&\underline{0.9962}	&\underline{68.6537}&0.2029				&1.2021\\
	\bb{SuperRPCA} 	    &\bb{0.9979}			&\bb{0.2645}				&\bb{0.0048}			&\bb{1.2624}				&\bb{0.9931}				&\bb{54.8237}				&\bb{0.2596}				&\bb{1.2575}			\\
	PTA		        &0.8676				&\underline{0.4063}	&0.1802				&1.2739				&0.6874				&2.2542				&0.2260				&1.0937\\
	Auto-AD	        &0.9259				&0.1114				&0.0079	            &1.0373				&0.9180				&14.0818			&0.1035				&1.0294\\
	RGAE	        &0.9795				&0.2401				&0.0166				&1.2197				&0.9630				&14.4938			&0.2236				&1.2031\\
	\bb{TAEF} 	    &\bb{0.9959}			&\bb{\textbf{0.4257}}				&\bb{0.0515}			&\bb{\textbf{1.4215}}				&\bb{0.9444}				&\bb{8.2726}				&\bb{\textbf{0.3742}}				&\bb{\textbf{1.3701}} \\
	Turbo-GoDec		&\textbf{0.9994}	&0.2001	            &\textbf{0.0027}	&1.1994	            &\textbf{0.9967}	&\textbf{74.1712}	&0.1974	            &1.1967\\
	\bottomrule
	\end{tabular}
\end{table*}

\subsubsection{Datasets} 
In the experiments, we selected three small-sized hyperspectral anomaly detection datasets: the HYDICE Urban dataset, the Pavia dataset, and the Hyperion dataset, with sizes of $80\times 100 \times 175$, $150 \times 150 \times 102$, and $100 \times 100 \times 145$ respectively.

\subsubsection{Comparison Methods} 
To fully demonstrate the effectiveness of Turbo-GoDec, we selected ten classic and effective anomaly detection algorithms as comparison methods, \dd{including GRX \cite{GRX}, LRX \cite{LRX}, CRD \cite{CRD}, GTVLRR \cite{GTVLRR}, LSMAD \cite{LSMAD}, SuperRPCA \cite{SuperRPCA}, PTA \cite{PTA}, Auto-AD \cite{Auto-AD}, RGAE \cite{RGAE}, and TAEF \cite{TAEF}. Among these, SuperRPCA and RGAE are methods that incorporate superpixel segmentation, while Auto-AD, RGAE, and TAEF are deep learning-based methods.
}

\subsubsection{Metrics}
To comprehensively evaluate and quantify the detection performance of Turbo-GoDec, we employed the 3D-ROC (three dimensional receiver operating characteristic) curve and its derived eight metrics to assess the overall effectiveness of the detection algorithms, as well as their anomaly detectability and background suppression capabilities \cite{3D-ROC-TD}, \cite{3D-ROC-AD}.
The eight derived metrics of 3D-ROC curve are
$\text{AUC}_\text{(D,F)}$, $\text{AUC}_{(\text{D},\tau)}$, $\text{AUC}_{(\text{F},\tau)}$, $\text{AUC}_\text{TD}$, $\text{AUC}_\text{BS}$, $\text{AUC}_\text{SNPR}$, $\text{AUC}_\text{TD-BS}$ and $\text{AUC}_\text{ODP}$. 

\subsubsection{Implementation Details}
To evaluate the generalization ability of Turbo-GoDec, we used the same set of parameters during the message passing on the factor graph. Specifically, the rank $r$ and the cardinality $k$ are based on NWHFC \cite{NWHFC} and MX-SVD \cite{MX-SVD}, \cite{MOCA}. 
During the message passing process, the message updates in the Markov random field are performed using smooth updates, which facilitate the convergence of the messages.
Moreover, the four parameters corresponding to the potential functions $\psi_{00}, \psi_{01}, \psi_{10}, \psi_{11}$ are set to 0.5, 0.3, 0.3, and 0.5, respectively. The trade-off parameter $\alpha$ is set to 0.4, 0.2 and 0.6 on the three datasets.\dd{The iteration number in S-step is set to 100.}

All experiments were conducted on a Windows 11 computer equipped with an Intel Core i7-13700K CPU and an NVIDIA RTX 4060 Ti GPU. \dd{Turbo-GoDec is implemented in MATLAB and does not use GPU acceleration. The code is available at https://github.com/jiahuisheng/Turbo-GoDec.}

\dd{\subsection{Ablation Study}}
Since our method primarily modifies the S-step of GoDec, the ablation version of Turbo-GoDec is the original GoDec, referred to as LSMAD in hyperspectral anomaly detection. Therefore, there is little need to conduct an additional ablation study here, as LSMAD has already been included in our comparative experiments. 

From Fig. \ref{fig:detection map} and Tables \ref{tab:HU_metric}, \ref{tab:Pavia_metric}, and \ref{tab:Hyperion_metric}, we can see that while LSMAD also demonstrates good background suppression capabilities, Turbo-GoDec shows even stronger background suppression, particularly evident in the Pavia dataset. Additionally, Turbo-GoDec outperforms LSMAD in the most metrics as well. The ablation study results can also be referred to in Section \ref{sec:Experiment on the non-cluster dataset}.

\subsection{Detection Performance}

In this section, we will explore the detection performance of Turbo-GoDec through the detection maps and the 3D-ROC derived metrics.
To demonstrate the effectiveness of Turbo-GoDec, the proposed Turbo-GoDec method is compared with other classical and effective anomaly detection methods on three real hyperspectral anomaly datasets. 
By examining the detection maps of each method, we can gain an intuitive sense of their detection performance. Meanwhile, the 3D-ROC derived metrics provide a quantitative assessment of the effectiveness of the methods. 
Based on the detection maps and the derived metrics, we will provide a detailed discussion of the detection performance across the three datasets.

\subsubsection{Performance over HYDICE Urban}
The detection maps over the HYDICE Urban dataset is presented in the first row of Fig. \ref{fig:detection map}. From the detection map, we can see that Turbo-GoDec successfully detected all anomalous pixels without any missed detections, and its background suppression capability is also commendable. Correspondingly, as shown in Table \ref{tab:HU_metric}, the metrics for Turbo-GoDec reflect this performance, with $\text{AUC}_{\text{(D,F)}}$ and $\text{AUC}_\text{SNPR}$ being second only to CRD, indicating a solid overall detection performance. Furthermore, the metrics representing background suppression capability, $\text{AUC}_{(\text{F},\tau)}$ and $\text{AUC}_{\text{BS}}$, are also second only to AutoAD, which aligns with the observations from the detection map.

\subsubsection{Performance over Pavia}
The detection maps over the Pavia dataset are shown in the second row of Fig. \ref{fig:detection map}. 
We can observe that certain methods, such as CRD, GTVLRR, PTA, and RGAE, encounter issues with poor background suppression when performing anomaly detection on the Pavia dataset. The bridges and land by the water show elevated anomalous scores, which indicates a higher incidence of false alarms.
In contrast, LSMAD, Auto-AD, and Turbo-GoDec demonstrate better background suppression capabilities.

The corresponding ROC curves and the derived metrics are presented in Fig. \ref{fig:Pavia_roc} and Table \ref{tab:Pavia_metric}. 
The metrics indicate that Turbo-GoDec achieves good detection performance while also demonstrating strong background suppression capabilities.

\subsubsection{Performance over Hyperion}
The anomaly detection maps over the Hyperion dataset is presented in the third row of Fig. \ref{fig:detection map}.
Similar to the previous two datasets, Turbo-GoDec also exhibits strong background suppression capabilities in the detection map for the Hyperion dataset.

The 3D-ROC curve and three 2D-ROC curves are displayed in Fig. \ref{fig:Hyperion_roc} and the ROC derived metrics are listed in Table \ref{tab:Hyperion_metric}. 
The metrics of $\text{AUC}_{\text{(D,F)}}$, $\text{AUC}_{(\text{F},\tau)}$, $\text{AUC}_{\text{BS}}$, and $\text{AUC}_{\text{SNPR}}$ demonstrate the outstanding overall detection capability and background suppression ability of Turbo-GoDec.

Overall, while Turbo-GoDec does not significantly outperform other methods in the metrics, the detection maps indicate that its overall detection performance is satisfactory, and its background suppression capability is also quite strong.



\dd{
\subsection{Parameter Analysis}

\begin{figure}[!h]
\includegraphics[width=\linewidth]{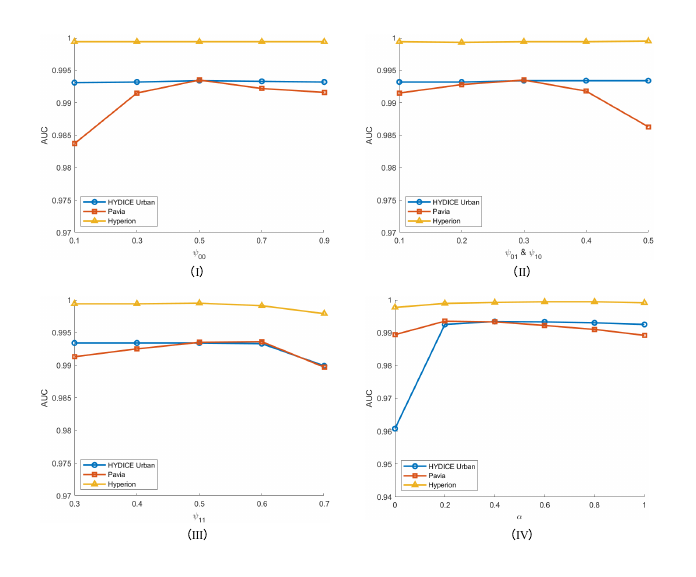}
\caption{
    \dd{
    The Impact of Key Parameters on Detection Performance ($\text{AUC}_{\text{(D,F)}}$).
    (\uppercase\expandafter{\romannumeral1}) Potential function $\psi_{00}$. 
    (\uppercase\expandafter{\romannumeral2}) Potential functions $\psi_{01}$ and $\psi_{10}$. 
    (\uppercase\expandafter{\romannumeral3}) Potential function $\psi_{11}$. 
    (\uppercase\expandafter{\romannumeral4}) The trade-off parameter $\alpha$.
    }
}
\label{fig:parameter_analysis}
\end{figure}

In this subsection, we conduct a parameter analysis of the key parameters in Turbo-GoDec, including the potential functions  $\psi_{00}, \psi_{01}, \psi_{10}, \psi_{11}$ in the Markov Random Field, as well as the trade-off parameter $\alpha$.

In the parameter analysis experiments, we fixed all other parameters and varied only the one being analyzed. As shown in Fig. \ref{fig:parameter_analysis}, overall, Turbo-GoDec is not very sensitive to the parameters $\psi_{00}, \psi_{01}, \psi_{10}$, and $\psi_{11}$ as long as they remain within a reasonable range. For the HYDICE Urban and Hyperion datasets, the $\text{AUC}_{\text{(D,F)}}$ variation with respect to $\psi_{00}$, $\psi_{01}$, and $\psi_{10}$ is within 0.003. 
In the Pavia dataset, the model exhibits relatively larger variations in performance metrics.: when $\psi_{00}$ is set to 0.1, the $\text{AUC}_{\text{(D,F)}}$ is 0.9837; when $\psi_{01}$ and $\psi_{10}$ are set to 0.5, the $\text{AUC}_{\text{(D,F)}}$ reaches 0.9863; and when $\psi_{11}$ is set to 0.7, the $\text{AUC}_{\text{(D,F)}}$ increases to 0.9897.
For the trade-off parameter $\alpha$, the model performance does not vary significantly when $\alpha$ is within the range of 0.2 to 0.6.

\subsection{Efficiency comparison}
In this subsection, we compare the runtime of different methods, including both non-DL-based and DL-based approaches. We calculate the runtime of each method on three datasets, and the results are presented in Table \ref{tab:runtime}.

\begin{table}[!h]
	\renewcommand{\arraystretch}{1.2}
	\centering
	\dd{
	\caption{Running times (in seconds) of the 11 methods on three hyperpsectral dataset.}
	\label{tab:runtime}
	\begin{tabular}{p{20mm}<{\centering}p{15mm}<{\centering}p{15mm}<{\centering}p{15mm}<{\centering}}
	\toprule
	$\text{Method}$ 	& $\text{HYDICE Urban}$ & $\text{Pavia}$ & $\text{Hyperion}$ \\\midrule
	GRX 	    &0.1174				&0.8325				&0.1548							\\
	LRX 	    &1.2218				&1.3031				&1.3038							\\
	CRD 	    &3.6516				&9.3552				&4.3849							\\
	GTVLRR 	    &58.9193			&137.5838			&70.4733							\\
	LSMAD 	    &6.1246				&11.1073			&6.3295				\\
	SuperRPCA   &5.0650				&5.8383				&3.8254							\\
	PTA 	    &6.9735				&11.2651			&6.9292							\\
	Auto-AD     &13.0863			&16.4474			&10.3038					\\
	RGAE 	    &64.0118			&307.8964			&84.5373							\\
	TAEF 	    &25.6036			&53.6162			&27.7678						\\
	Turbo-GoDec	&6.1868				&13.8326			&6.7162	         	\\
	\bottomrule
	\end{tabular}
	}
\end{table}

Overall, non-DL-based methods are less time-consuming than DL-based methods, with GRX having the shortest runtime.Meanwhile, we found that our Turbo-GoDec does not exhibit a significant increase in runtime compared to LSMAD. This is because the main time consumption lies in the SVD operation in the L-step. The time complexity of the L-step is $O(HWC^2)$, while that of the S-step is 
$O(HW)$.

It is important to note that these methods are implemented on different hardware and software platforms. Thus, the comparison may not be strictly fair. Among all the methods, only Auto-AD employs GPU acceleration.

\subsection{Experiment on the non-cluster dataset \& Limitation}
\label{sec:Experiment on the non-cluster dataset}

\subsubsection{Experiment on the non-cluster dataset}
Since Turbo-GoDec is based on the cluster sparsity assumption, it may theoretically suppress the anomaly scores of some dispersed and large-scale anomalies. To investigate this issue, we conducted additional experiments on the three real datasets and the synthetic Salinas dataset, which contains dispersed anomalies. 

The detection maps and corresponding metrics are presented in Fig. \ref{fig:sparsity_experiment} and Table \ref{tab:sparsity_experiment}. In Fig. \ref{fig:sparsity_experiment}, the first two columns show the detection results of Turbo-GoDec without the cluster sparsity prior (CSP) constraint, while the last two columns show the results with the cluster sparsity prior constraint. From the detection maps of the first three rows and the metrics in Table \ref{tab:sparsity_experiment}, we can see that for the three real datasets that satisfy the cluster sparsity assumption, incorporating the prior effectively highlights the anomalies while better suppressing the background. Taking the Pavia dataset as an example, the $\text{AUC}_{\text{(D,F)}}$ improves from 0.9102 to 0.9935, and the $\text{AUC}_{(\text{F},\tau)}$ decreases from 0.0187 to 0.0011. In the last row of Fig. \ref{fig:sparsity_experiment}, Turbo-GoDec shows a different result on the non-cluster sparsity dataset Salinas. As highlighted by the red box in the figure, the cluster sparsity prior constraint further weakens the anomaly saliency of originally weak and dispersed anomalies, leading to some missed detections. Correspondingly, the $\text{AUC}_{\text{(D,F)}}$ drops from 0.9979 to 0.9444.

\begin{figure}[!h]
\includegraphics[width=0.98\linewidth]{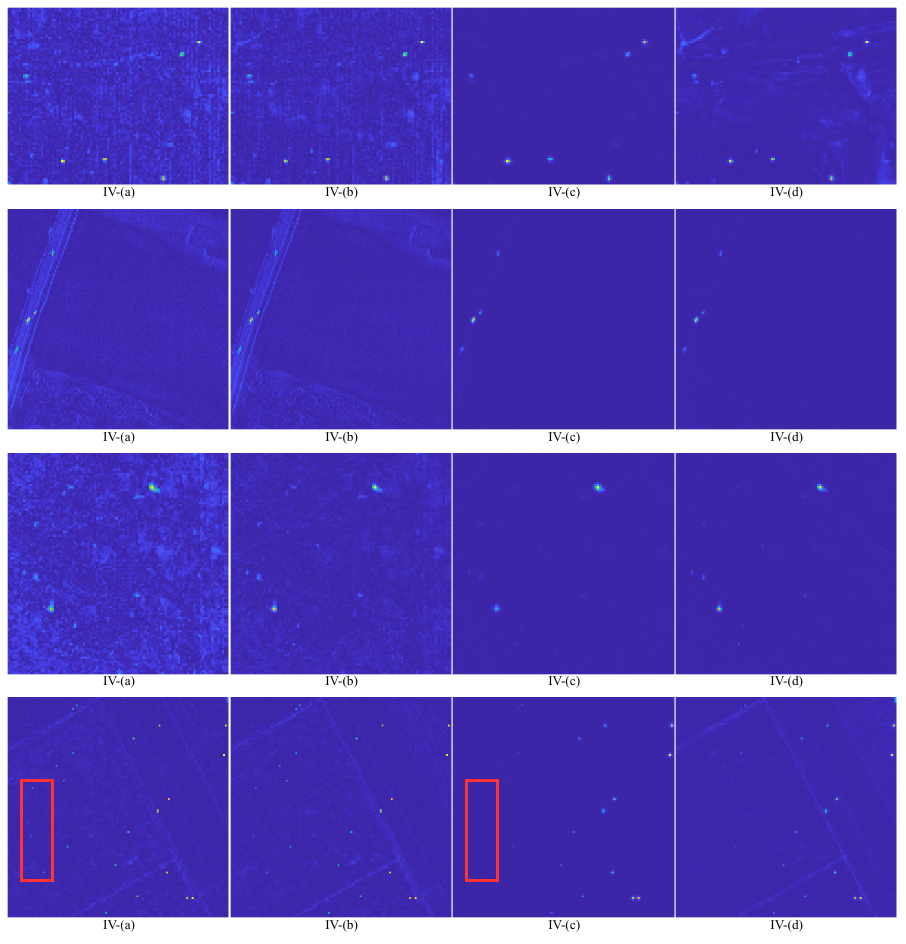}
\caption{
    \dd{
    The detection map of the ablation experiment on real datasets and non-clustered synthetic dataset (Salinas).
    (\uppercase\expandafter{\romannumeral1}) HYDICE Urban. 
    (\uppercase\expandafter{\romannumeral2}) Pavia. 
    (\uppercase\expandafter{\romannumeral3}) Hyperion. 
    (\uppercase\expandafter{\romannumeral4}) Salinas.
    (a) detection map of $T$ in Turbo-GoDec without cluster sparsity piror constraint.
    (b) detection map of Turbo-GoDec without cluster sparsity piror constraint.
    (c) detection map of $J$ in Turbo-GoDec.
    (c) detection map of Turbo-GoDec.
    }
}
\label{fig:sparsity_experiment}
\end{figure}

\begin{table}[!h]
	\renewcommand{\arraystretch}{1.2}
	\centering
	\dd{
	\caption{Ablation study results of the cluster sparsity prior (CSP) constraint on both real datasets and synthetic dataset (non-clustered dataset).}
	\label{tab:sparsity_experiment}
	\begin{tabular}{p{10mm}<{\centering}p{10mm}<{\centering}p{10mm}<{\centering}p{10mm}<{\centering}p{10mm}<{\centering}p{10mm}<{\centering}}
	\toprule
	$\text{Metric}$ 	& $\text{CSP}$ 	& $\text{Urban}$ & $\text{Pavia}$ & $\text{Hyperion}$ & $\text{Salinas}$\\\midrule
    \multirow{2}{*}{$\text{AUC}_{\text{(D,F)}}$}    &\cmark 	    &\textbf{0.9934}	&\textbf{0.9935}	&\textbf{0.9994}&0.9444				\\
	                                                &\xmark 	    &0.9836				&0.9102				&0.9993			&\textbf{0.9979}	\\\midrule
	\multirow{2}{*}{$\text{AUC}_{(\text{F},\tau)}$} &\cmark 	    &\textbf{0.0145}	&\textbf{0.0011}	&\textbf{0.0027}&\textbf{0.0094}	\\
	                                                &\xmark 	    &0.0272				&0.0187				&0.0176			&0.0112				\\
	\bottomrule
	\end{tabular}
	}
\end{table}

\subsubsection{Limitation}
In Turbo-GoDec, we assume that anomalies in hyperspectral images exhibit the characteristic of cluster sparsity. Although this assumption holds true for most scenarios, there are cases where it does not apply. For instance, in our synthetic dataset (Salinas), the anomalies are dispersed and do not conform to the cluster sparsity assumption. Similarly, in the San Diego dataset, the airplane targets also challenge the suitability of the cluster sparsity assumption (and may even violate the general sparsity assumption). However, due to ongoing debate among researchers regarding whether large-scale targets should be classified as anomalies, we do not further discuss large-scale anomalies dataset in this work.

\subsection{Discussion}
Turbo-GoDec actively leverages the cluster sparsity prior, whereas other methods that utilize spatial information of anomalies do so passively. For example, LRX \cite{LRX} takes into account the possibility that the neighboring pixels of the test pixel may be contaminated by anomalies, and thus adopts a dual-window strategy to obtain a cleaner background (it does not actively exploit cluster sparsity prior). Similarly, superpixel-based methods, as the one described in \cite{Superpixel-ref-1}, use SLIC \cite{SLIC} to remove small regions that might be anomalies. These methods all rely on the fact that anomalies tend to occupy a certain spatial extent, and use this prior knowledge to avoid the contamination of the background by anomalies rather than actively leverage this prior knowledge to detect anomalies. This passive strategy is even more apparent in \cite{SPDWRX}, which combines superpixel segmentation with LRX, using superpixel segmentation to precisely remove potential anomalous contamination in the neighborhood.

All of these methods consider the potential negative impact of cluster sparsity and aim to avoid it. In contrast, our method actively exploits the cluster sparsity prior by guiding the model to search for anomalous regions with a certain spatial size.

}

\section{Conclusion}
In this paper, we proposed a new anomaly detection method based on the GoDec model, called Turbo-GoDec, which considers the spatial cluster sparsity prior of anomalous targets. 
First, we modeled the cluster sparsity prior using a Markov random field and transformed it into a factor graph format. Next, we implemented message passing on the factor graph to compute the marginal probabilities of each pixel being anomalous. Subsequently, we updated the sparse component in GoDec based on the obtained marginal probabilities. Finally, we processed the resulting low-rank component along with the marginal probabilities to generate our anomaly detection map.
We also conducted experiments to validate the effectiveness of Turbo-GoDec. 

Following the framework of Turbo-GoDec, the cluster sparsity prior can similarly be applied to other matrix decomposition-based anomaly detection methods, such as RPCA \cite{RPCA}, \cite{RPCA-AD}. We hope that Turbo-GoDec can provide valuable insights for others.

\footnotesize
\bibliographystyle{IEEEtran}
\bibliography{ref}

@ARTICLE{SLIC,
  author={Achanta, Radhakrishna and Shaji, Appu and Smith, Kevin and Lucchi, Aurelien and Fua, Pascal and Süsstrunk, Sabine},
  journal={IEEE Transactions on Pattern Analysis and Machine Intelligence}, 
  title={SLIC Superpixels Compared to State-of-the-Art Superpixel Methods}, 
  year={2012},
  volume={34},
  number={11},
  pages={2274-2282},
  doi={10.1109/TPAMI.2012.120}}

@ARTICLE{SPDWRX,
  author={Ren, Lang and Zhao, Liaoying and Wang, Yulei},
  journal={IEEE Geoscience and Remote Sensing Letters}, 
  title={A Superpixel-Based Dual Window RX for Hyperspectral Anomaly Detection}, 
  year={2020},
  volume={17},
  number={7},
  pages={1233-1237},
  doi={10.1109/LGRS.2019.2942949}}

@INPROCEEDINGS{Superpixel-ref-1,
  author={Arisoy, Sertac and Kayabol, Koray},
  booktitle={2019 10th Workshop on Hyperspectral Imaging and Signal Processing: Evolution in Remote Sensing (WHISPERS)}, 
  title={Anomaly Detection In Hyperspectral Images Via Superpixel Segmentation And Unsupervised Background Learning}, 
  year={2019},
  volume={},
  number={},
  pages={1-5},
  doi={10.1109/WHISPERS.2019.8921261}}

@ARTICLE{KIFD,
  author={Li, Shutao and Zhang, Kunzhong and Duan, Puhong and Kang, Xudong},
  journal={IEEE Transactions on Geoscience and Remote Sensing}, 
  title={Hyperspectral Anomaly Detection With Kernel Isolation Forest}, 
  year={2020},
  volume={58},
  number={1},
  pages={319-329},
  doi={10.1109/TGRS.2019.2936308}}

@ARTICLE{doCSVM,
  author={Miao, Xuedan and Liu, Ying and Zhao, Haiquan and Li, Chunguang},
  journal={IEEE Transactions on Cybernetics}, 
  title={Distributed Online One-Class Support Vector Machine for Anomaly Detection Over Networks}, 
  year={2019},
  volume={49},
  number={4},
  pages={1475-1488},
  doi={10.1109/TCYB.2018.2804940}}

@ARTICLE{2S-GLRT,
  author={Liu, Jun and Hou, Zengfu and Li, Wei and Tao, Ran and Orlando, Danilo and Li, Hongbin},
  journal={IEEE Transactions on Neural Networks and Learning Systems}, 
  title={Multipixel Anomaly Detection With Unknown Patterns for Hyperspectral Imagery}, 
  year={2022},
  volume={33},
  number={10},
  pages={5557-5567},
  doi={10.1109/TNNLS.2021.3071026}}

@ARTICLE{SuperRPCA,
  author={Lin, Jhao-Ting and Lin, Chia-Hsiang},
  journal={IEEE Transactions on Geoscience and Remote Sensing}, 
  title={SuperRPCA: A Collaborative Superpixel Representation Prior-Aided RPCA for Hyperspectral Anomaly Detection}, 
  year={2024},
  volume={62},
  number={},
  pages={1-16},
  doi={10.1109/TGRS.2024.3459956}}

@ARTICLE{PSSIN,
  author={Cheng, Xi and Wang, Chenhao and Huo, Yu and Zhang, Min and Wang, Hai and Ren, Jinchang},
  journal={IEEE Transactions on Geoscience and Remote Sensing}, 
  title={Prototype-Guided Spatial-Spectral Interaction Network for Hyperspectral Anomaly Detection}, 
  year={2025},
  volume={},
  number={},
  pages={1-1},
  doi={10.1109/TGRS.2025.3568121}}

@ARTICLE{Memory-Augmented-AE,
  author={Huo, Yu and Cheng, Xi and Lin, Sheng and Zhang, Min and Wang, Hai},
  journal={IEEE Transactions on Geoscience and Remote Sensing}, 
  title={Memory-Augmented Autoencoder With Adaptive Reconstruction and Sample Attribution Mining for Hyperspectral Anomaly Detection}, 
  year={2024},
  volume={62},
  number={},
  pages={1-18},
  doi={10.1109/TGRS.2024.3399313}}

@ARTICLE{WRX,
  author={Guo, Qiandong and Zhang, Bing and Ran, Qiong and Gao, Lianru and Li, Jun and Plaza, Antonio},
  journal={IEEE Journal of Selected Topics in Applied Earth Observations and Remote Sensing}, 
  title={Weighted-RXD and Linear Filter-Based RXD: Improving Background Statistics Estimation for Anomaly Detection in Hyperspectral Imagery}, 
  year={2014},
  volume={7},
  number={6},
  pages={2351-2366},
  doi={10.1109/JSTARS.2014.2302446}}

@ARTICLE{FrFT-RC-LRaSMD,
  author={Ma, Yong and Fan, Ganghui and Jin, Qiwen and Huang, Jun and Mei, Xiaoguang and Ma, Jiayi},
  journal={IEEE Geoscience and Remote Sensing Letters}, 
  title={Hyperspectral Anomaly Detection via Integration of Feature Extraction and Background Purification}, 
  year={2021},
  volume={18},
  number={8},
  pages={1436-1440},
  doi={10.1109/LGRS.2020.2998809}}

@article{TRPCA-AEAN,
author = {Emoto, Atsuya and Matsuoka, Ryo},
year = {2025},
month = {01},
pages = {1-1},
title = {Unsupervised Anomaly Detection in Hyperspectral Imaging: Integrating Tensor Robust Principal Component Analysis With Autoencoding Adversarial Networks},
volume = {13},
journal = {IEEE Access},
doi = {10.1109/ACCESS.2025.3534981}
}

@INPROCEEDINGS{LTNN,
  author={Mishima, Mio and Kobayashi, Iori and Matsuoka, Ryo},
  booktitle={IGARSS 2023 - 2023 IEEE International Geoscience and Remote Sensing Symposium}, 
  title={Hyperspectral Anomaly Detection Based on Local-Tensor-Nuclear-Norm}, 
  year={2023},
  volume={},
  number={},
  pages={2157-2160},
  doi={10.1109/IGARSS52108.2023.10281637}}

@ARTICLE{WeaklyAD,
  author={Jiang, Tao and Xie, Weiying and Li, Yunsong and Lei, Jie and Du, Qian},
  journal={IEEE Transactions on Neural Networks and Learning Systems}, 
  title={Weakly Supervised Discriminative Learning With Spectral Constrained Generative Adversarial Network for Hyperspectral Anomaly Detection}, 
  year={2022},
  volume={33},
  number={11},
  pages={6504-6517},
  doi={10.1109/TNNLS.2021.3082158}}

@ARTICLE{TAEF,
  author={Wu, Ziyu and Wang, Bin},
  journal={IEEE Transactions on Geoscience and Remote Sensing}, 
  title={Transformer-Based Autoencoder Framework for Nonlinear Hyperspectral Anomaly Detection}, 
  year={2024},
  volume={62},
  number={},
  pages={1-15},
  doi={10.1109/TGRS.2024.3361469}}

@ARTICLE{GT-HAD,
  author={Lian, Jie and Wang, Lizhi and Sun, He and Huang, Hua},
  journal={IEEE Transactions on Neural Networks and Learning Systems}, 
  title={GT-HAD: Gated Transformer for Hyperspectral Anomaly Detection}, 
  year={2025},
  volume={36},
  number={2},
  pages={3631-3645},
  doi={10.1109/TNNLS.2024.3355166}}

@ARTICLE{DWSDiff,
  author={Chen, Wenbin and Zhi, Xiyang and Jiang, Shikai and Huang, Yuanxin and Han, Qichao and Zhang, Wei},
  journal={IEEE Transactions on Geoscience and Remote Sensing}, 
  title={DWSDiff: Dual-Window Spectral Diffusion for Hyperspectral Anomaly Detection}, 
  year={2025},
  volume={63},
  number={},
  pages={1-17},
  doi={10.1109/TGRS.2025.3537331}}

@misc{BSDM,
      title={BSDM: Background Suppression Diffusion Model for Hyperspectral Anomaly Detection}, 
      author={Jitao Ma and Weiying Xie and Yunsong Li and Leyuan Fang},
      year={2023},
      eprint={2307.09861},
      archivePrefix={arXiv},
      primaryClass={cs.CV},
      url={https://arxiv.org/abs/2307.09861}, 
}

@ARTICLE{PCA-TLRSR,
  author={Wang, Minghua and Wang, Qiang and Hong, Danfeng and Roy, Swalpa Kumar and Chanussot, Jocelyn},
  journal={IEEE Transactions on Cybernetics}, 
  title={Learning Tensor Low-Rank Representation for Hyperspectral Anomaly Detection}, 
  year={2023},
  volume={53},
  number={1},
  pages={679-691},
  doi={10.1109/TCYB.2022.3175771}}

@ARTICLE{CGCRD,
  author={Zhao, Xiaobin and Li, Wei and Zhao, Chunhui and Tao, Ran},
  journal={IEEE Transactions on Geoscience and Remote Sensing}, 
  title={Hyperspectral Target Detection Based on Weighted Cauchy Distance Graph and Local Adaptive Collaborative Representation}, 
  year={2022},
  volume={60},
  number={},
  pages={1-13},
  doi={10.1109/TGRS.2022.3169171}}

@ARTICLE{SPSTT,
  author={Zhao, Xiaobin and Liu, Kaiqi and Gao, Kun and Li, Wei},
  journal={IEEE Transactions on Geoscience and Remote Sensing}, 
  title={Hyperspectral Time-Series Target Detection Based on Spectral Perception and Spatial–Temporal Tensor Decomposition}, 
  year={2023},
  volume={61},
  number={},
  pages={1-12},
  doi={10.1109/TGRS.2023.3307071}}

@ARTICLE{TRGLF,
  author={Zhao, Xiaobin and Liu, Kaiqi and Wang, Xueying and Zhao, Song and Gao, Kun and Lin, Hongyang and Zong, Yantao and Li, Wei},
  journal={IEEE Journal of Selected Topics in Applied Earth Observations and Remote Sensing}, 
  title={Tensor Adaptive Reconstruction Cascaded With Global and Local Feature Fusion for Hyperspectral Target Detection}, 
  year={2025},
  volume={18},
  number={},
  pages={607-620},
  doi={10.1109/JSTARS.2024.3486774}}

@inproceedings{BSN,
  title={Noise2void-learning denoising from single noisy images},
  author={Krull, Alexander and Buchholz, Tim-Oliver and Jug, Florian},
  booktitle={Proceedings of the IEEE/CVF conference on computer vision and pattern recognition},
  pages={2129--2137},
  year={2019}
}

@ARTICLE{NWHFC,
  author={Chein-I Chang and Qian Du},
  journal={IEEE Transactions on Geoscience and Remote Sensing}, 
  title={Estimation of number of spectrally distinct signal sources in hyperspectral imagery}, 
  year={2004},
  volume={42},
  number={3},
  pages={608-619},
  doi={10.1109/TGRS.2003.819189}}

@article{MOCA,
  title={Rank estimation and redundancy reduction of high-dimensional noisy signals with preservation of rare vectors},
  author={Kuybeda, Oleg and Malah, David and Barzohar, Meir},
  journal={IEEE Transactions on Signal Processing},
  volume={55},
  number={12},
  pages={5579--5592},
  year={2007},
  publisher={IEEE}
}

@ARTICLE{MX-SVD,
  author={Chang, Chein-I and Cao, Hongju and Chen, Shuhan and Shang, Xiaodi and Yu, Chunyan and Song, Meiping},
  journal={IEEE Transactions on Geoscience and Remote Sensing}, 
  title={Orthogonal Subspace Projection-Based Go-Decomposition Approach to Finding Low-Rank and Sparsity Matrices for Hyperspectral Anomaly Detection}, 
  year={2021},
  volume={59},
  number={3},
  pages={2403-2429},
  doi={10.1109/TGRS.2020.3002724}}

@article{RPCA,
  title={Robust uncertainty principles: Exact signal reconstruction from highly incomplete frequency information},
  author={Cand{\`e}s, Emmanuel J and Romberg, Justin and Tao, Terence},
  journal={IEEE Transactions on information theory},
  volume={52},
  number={2},
  pages={489--509},
  year={2006},
  publisher={IEEE}
}

@inproceedings{RPCA-AD,
  title={Low-rank decomposition-based anomaly detection},
  author={Chen, Shih-Yu and Yang, Shiming and Kalpakis, Konstantinos and Chang, Chein-I},
  booktitle={Algorithms and Technologies for Multispectral, Hyperspectral, and Ultraspectral Imagery XIX},
  volume={8743},
  pages={171--177},
  year={2013},
  organization={SPIE}
}

@ARTICLE{Sparsity_model,
  author={Huang, Zhe and Wang, Kexuan and Liu, An and Cai, Yunlong and Du, Rui and Han, Tony Xiao},
  journal={IEEE Transactions on Wireless Communications}, 
  title={Joint Pilot Optimization, Target Detection and Channel Estimation for Integrated Sensing and Communication Systems}, 
  year={2022},
  volume={21},
  number={12},
  pages={10351-10365},
  doi={10.1109/TWC.2022.3183621}}

@ARTICLE{Sparsity_model_2,
  author={Lian, Lixiang and Liu, An and Lau, Vincent K. N.},
  journal={IEEE Transactions on Signal Processing}, 
  title={Exploiting Dynamic Sparsity for Downlink FDD-Massive MIMO Channel Tracking}, 
  year={2019},
  volume={67},
  number={8},
  pages={2007-2021},
  doi={10.1109/TSP.2019.2896179}}

@ARTICLE{Turbo-CS,
  author={Ma, Junjie and Yuan, Xiaojun and Ping, Li},
  journal={IEEE Signal Processing Letters}, 
  title={On the Performance of Turbo Signal Recovery with Partial DFT Sensing Matrices}, 
  year={2015},
  volume={22},
  number={10},
  pages={1580-1584},
  doi={10.1109/LSP.2015.2414951}}

@book{MRF,
  title={Markov random field modeling in image analysis},
  author={Li, Stan Z},
  year={2009},
  publisher={Springer Science \& Business Media}
}

@inproceedings{grid-MRF,
author = {Petersen, Kersten and Burkhardt, Hans},
year = {2008},
month = {06},
pages = {41-50},
title = {Fast Generalized Belief Propagation for MAP Estimation on 2D and 3D Grid-Like Markov Random Fields},
isbn = {978-3-540-69320-8},
doi = {10.1007/978-3-540-69321-5_5}
}

@ARTICLE{factor-graph,
  author={Kschischang, F.R. and Frey, B.J. and Loeliger, H.-A.},
  journal={IEEE Transactions on Information Theory}, 
  title={Factor graphs and the sum-product algorithm}, 
  year={2001},
  volume={47},
  number={2},
  pages={498-519},
  doi={10.1109/18.910572}}

@inproceedings{GoDec,
  title={Godec: Randomized low-rank \& sparse matrix decomposition in noisy case},
  author={Zhou, Tianyi and Tao, Dacheng},
  booktitle={Proceedings of the 28th International Conference on Machine Learning, ICML 2011},
  year={2011}
}

@ARTICLE{background-1,
  author={Bioucas-Dias, Jose M. and Plaza, Antonio and Camps-Valls, Gustavo and Scheunders, Paul and Nasrabadi, Nasser and Chanussot, Jocelyn},
  journal={IEEE Geoscience and Remote Sensing Magazine}, 
  title={Hyperspectral Remote Sensing Data Analysis and Future Challenges}, 
  year={2013},
  volume={1},
  number={2},
  pages={6-36},
  doi={10.1109/MGRS.2013.2244672}}

@ARTICLE{background-2,
  author={Ghamisi, Pedram and Yokoya, Naoto and Li, Jun and Liao, Wenzhi and Liu, Sicong and Plaza, Javier and Rasti, Behnood and Plaza, Antonio},
  journal={IEEE Geoscience and Remote Sensing Magazine}, 
  title={Advances in Hyperspectral Image and Signal Processing: A Comprehensive Overview of the State of the Art}, 
  year={2017},
  volume={5},
  number={4},
  pages={37-78},
  doi={10.1109/MGRS.2017.2762087}}

@ARTICLE{usage-1,
  author={Chein-I Chang and Shao-Shan Chiang},
  journal={IEEE Transactions on Geoscience and Remote Sensing}, 
  title={Anomaly detection and classification for hyperspectral imagery}, 
  year={2002},
  volume={40},
  number={6},
  pages={1314-1325},
  doi={10.1109/TGRS.2002.800280}}

@ARTICLE{usage-2,
  author={Nasrabadi, Nasser M.},
  journal={IEEE Signal Processing Magazine}, 
  title={Hyperspectral Target Detection : An Overview of Current and Future Challenges}, 
  year={2014},
  volume={31},
  number={1},
  pages={34-44},
  doi={10.1109/MSP.2013.2278992}}

@ARTICLE{usage-TD,
  author={Zhu, Dehui and Du, Bo and Dong, Yanni and Zhang, Liangpei},
  journal={IEEE Transactions on Multimedia}, 
  title={Target Detection With Spatial-Spectral Adaptive Sample Generation and Deep Metric Learning for Hyperspectral Imagery}, 
  year={2023},
  volume={25},
  number={},
  pages={6538-6550},
  doi={10.1109/TMM.2022.3210389}}

@ARTICLE{usage-Classification,
  author={Shi, Cheng and Pun, Chi-Man},
  journal={IEEE Transactions on Multimedia}, 
  title={Multiscale Superpixel-Based Hyperspectral Image Classification Using Recurrent Neural Networks With Stacked Autoencoders}, 
  year={2020},
  volume={22},
  number={2},
  pages={487-501},
  doi={10.1109/TMM.2019.2928491}}

@ARTICLE{anomaly-1,
  author={Matteoli, Stefania and Diani, Marco and Corsini, Giovanni},
  journal={IEEE Aerospace and Electronic Systems Magazine}, 
  title={A tutorial overview of anomaly detection in hyperspectral images}, 
  year={2010},
  volume={25},
  number={7},
  pages={5-28},
  doi={10.1109/MAES.2010.5546306}}

@ARTICLE{anomaly-2,
  author={Chang, Chein-I.},
  journal={IEEE Transactions on Geoscience and Remote Sensing}, 
  title={Hyperspectral Anomaly Detection: A Dual Theory of Hyperspectral Target Detection}, 
  year={2022},
  volume={60},
  number={},
  pages={1-20},
  doi={10.1109/TGRS.2021.3086768}}

@ARTICLE{anomaly-def-1,
  author={{C.-I. Chang}},
  journal={IEEE Transactions on Geoscience and Remote Sensing}, 
  title={Effective Anomaly Space for Hyperspectral Anomaly Detection}, 
  year={2022},
  volume={60},
  number={},
  pages={1-24},
  doi={10.1109/TGRS.2022.3161632}}

@ARTICLE{LRX,
  author={Molero, José Manuel and Garzón, Ester M. and García, Inmaculada and Plaza, Antonio},
  journal={IEEE Journal of Selected Topics in Applied Earth Observations and Remote Sensing}, 
  title={Analysis and Optimizations of Global and Local Versions of the RX Algorithm for Anomaly Detection in Hyperspectral Data}, 
  year={2013},
  volume={6},
  number={2},
  pages={801-814},
  doi={10.1109/JSTARS.2013.2238609}}

@ARTICLE{KRX,
  author={Heesung Kwon and Nasrabadi, N.M.},
  journal={IEEE Transactions on Geoscience and Remote Sensing}, 
  title={Kernel RX-algorithm: a nonlinear anomaly detector for hyperspectral imagery}, 
  year={2005},
  volume={43},
  number={2},
  pages={388-397},
  doi={10.1109/TGRS.2004.841487}}

@ARTICLE{FrRX,
  author={Tao, Ran and Zhao, Xudong and Li, Wei and Li, Heng-Chao and Du, Qian},
  journal={IEEE Journal of Selected Topics in Applied Earth Observations and Remote Sensing}, 
  title={Hyperspectral Anomaly Detection by Fractional Fourier Entropy}, 
  year={2019},
  volume={12},
  number={12},
  pages={4920-4929},
  doi={10.1109/JSTARS.2019.2940278}}

@ARTICLE{LRASR,
  author={Xu, Yang and Wu, Zebin and Li, Jun and Plaza, Antonio and Wei, Zhihui},
  journal={IEEE Transactions on Geoscience and Remote Sensing}, 
  title={Anomaly Detection in Hyperspectral Images Based on Low-Rank and Sparse Representation}, 
  year={2016},
  volume={54},
  number={4},
  pages={1990-2000},
  doi={10.1109/TGRS.2015.2493201}}

@ARTICLE{SD_LRASR,
  author={Zhang, Xiangrong and Ma, Xiaoxiao and Huyan, Ning and Gu, Jing and Tang, Xu and Jiao, Licheng},
  journal={IEEE Transactions on Geoscience and Remote Sensing}, 
  title={Spectral-Difference Low-Rank Representation Learning for Hyperspectral Anomaly Detection}, 
  year={2021},
  volume={59},
  number={12},
  pages={10364-10377},
  doi={10.1109/TGRS.2020.3046727}}

@ARTICLE{LRASR_BP,
  author={Huyan, Ning and Zhang, Xiangrong and Zhou, Huiyu and Jiao, Licheng},
  journal={IEEE Transactions on Geoscience and Remote Sensing}, 
  title={Hyperspectral Anomaly Detection via Background and Potential Anomaly Dictionaries Construction}, 
  year={2019},
  volume={57},
  number={4},
  pages={2263-2276},
  doi={10.1109/TGRS.2018.2872590}}

@ARTICLE{LRASR_UD,
  author={Qu, Ying and Wang, Wei and Guo, Rui and Ayhan, Bulent and Kwan, Chiman and Vance, Steven and Qi, Hairong},
  journal={IEEE Transactions on Geoscience and Remote Sensing}, 
  title={Hyperspectral Anomaly Detection Through Spectral Unmixing and Dictionary-Based Low-Rank Decomposition}, 
  year={2018},
  volume={56},
  number={8},
  pages={4391-4405},
  doi={10.1109/TGRS.2018.2818159}}

@INPROCEEDINGS{CR_PCA,
  author={Wu, Zhaoyue and Su, Hongiun and Zheng, Pan},
  booktitle={2018 9th Workshop on Hyperspectral Image and Signal Processing: Evolution in Remote Sensing (WHISPERS)}, 
  title={Hyperspectral Anomaly Detection Using Collaborative Representation with PCA Remove Outlier}, 
  year={2018},
  volume={},
  number={},
  pages={1-5},
  doi={10.1109/WHISPERS.2018.8747083}}

@ARTICLE{RCRD,
  author={Wu, Zhaoyue and Su, Hongjun and Tao, Xuanwen and Han, Lirong and Paoletti, Mercedes E. and Haut, Juan M. and Plaza, Javier and Plaza, Antonio},
  journal={IEEE Transactions on Geoscience and Remote Sensing}, 
  title={Hyperspectral Anomaly Detection With Relaxed Collaborative Representation}, 
  year={2022},
  volume={60},
  number={},
  pages={1-17},
  doi={10.1109/TGRS.2022.3190327}}

@article{morCRD,
author = {Maryam Imani},
title = {Anomaly detection using morphology-based collaborative representation in hyperspectral imagery},
journal = {European Journal of Remote Sensing},
volume = {51},
number = {1},
pages = {457--471},
year = {2018},
publisher = {Taylor \& Francis},
doi = {10.1080/22797254.2018.1446727},
}

@article{swCRD,
title = {Self-weighted collaborative representation for hyperspectral anomaly detection},
journal = {Signal Processing},
volume = {177},
pages = {107718},
year = {2020},
issn = {0165-1684},
doi = {https://doi.org/10.1016/j.sigpro.2020.107718},
author = {Rong Wang and Haojie Hu and Fang He and Feiping Nie and Shubin Cai and Zhong Ming}
}

@article{LRaSMD,
author = {Weiwei Sun and Chun Liu and Jialin Li and Yenming Mark Lai and Weiyue Li},
title = {Low-rank and sparse matrix decomposition-based anomaly detection for hyperspectral imagery},
volume = {8},
journal = {Journal of Applied Remote Sensing},
number = {1},
publisher = {SPIE},
pages = {083641},
year = {2014},
doi = {10.1117/1.JRS.8.083641},
}

@ARTICLE{LSDM-MoG,
  author={Li, Lu and Li, Wei and Du, Qian and Tao, Ran},
  journal={IEEE Transactions on Cybernetics}, 
  title={Low-Rank and Sparse Decomposition With Mixture of Gaussian for Hyperspectral Anomaly Detection}, 
  year={2021},
  volume={51},
  number={9},
  pages={4363-4372},
  doi={10.1109/TCYB.2020.2968750}}

@ARTICLE{Tucker,
  author={Zhang, Xing and Wen, Gongjian and Dai, Wei},
  journal={IEEE Transactions on Geoscience and Remote Sensing}, 
  title={A Tensor Decomposition-Based Anomaly Detection Algorithm for Hyperspectral Image}, 
  year={2016},
  volume={54},
  number={10},
  pages={5801-5820},
  doi={10.1109/TGRS.2016.2572400}}

@ARTICLE{FA_TD,
  author={Zhang, Xing and Wen, Gongjian},
  journal={IEEE Geoscience and Remote Sensing Letters}, 
  title={A Fast and Adaptive Method for Determining $K_{1}$ , $K_{2}$ , and $K_{3}$ in the Tensor Decomposition-Based Anomaly Detection Algorithm}, 
  year={2018},
  volume={15},
  number={1},
  pages={3-7},
  doi={10.1109/LGRS.2017.2759963}}

@ARTICLE{PBTA,
  author={Li, Lu and Li, Wei and Qu, Ying and Zhao, Chunhui and Tao, Ran and Du, Qian},
  journal={IEEE Transactions on Neural Networks and Learning Systems}, 
  title={Prior-Based Tensor Approximation for Anomaly Detection in Hyperspectral Imagery}, 
  year={2022},
  volume={33},
  number={3},
  pages={1037-1050},
  doi={10.1109/TNNLS.2020.3038659}}

@ARTICLE{TensorRPCA,
  author={A, Ruhan and Mu, Xiaodong and He, Jingyuan},
  journal={IEEE Geoscience and Remote Sensing Letters}, 
  title={Enhance Tensor RPCA-Based Mahalanobis Distance Method for Hyperspectral Anomaly Detection}, 
  year={2022},
  volume={19},
  number={},
  pages={1-5},
  doi={10.1109/LGRS.2022.3159341}}

@ARTICLE{BS3LNet,
  author={Gao, Lianru and Wang, Degang and Zhuang, Lina and Sun, Xu and Huang, Min and Plaza, Antonio},
  journal={IEEE Transactions on Geoscience and Remote Sensing}, 
  title={BS3LNet: A New Blind-Spot Self-Supervised Learning Network for Hyperspectral Anomaly Detection}, 
  year={2023},
  volume={61},
  number={},
  pages={1-18},
  doi={10.1109/TGRS.2023.3246565}}

@ARTICLE{PDBSNet,
  author={Wang, Degang and Zhuang, Lina and Gao, Lianru and Sun, Xu and Huang, Min and Plaza, Antonio J.},
  journal={IEEE Transactions on Geoscience and Remote Sensing}, 
  title={PDBSNet: Pixel-Shuffle Downsampling Blind-Spot Reconstruction Network for Hyperspectral Anomaly Detection}, 
  year={2023},
  volume={61},
  number={},
  pages={1-14},
  doi={10.1109/TGRS.2023.3276175}}

@ARTICLE{CNN-based,
  author={Li, Wei and Wu, Guodong and Du, Qian},
  journal={IEEE Geoscience and Remote Sensing Letters}, 
  title={Transferred Deep Learning for Anomaly Detection in Hyperspectral Imagery}, 
  year={2017},
  volume={14},
  number={5},
  pages={597-601},
  doi={10.1109/LGRS.2017.2657818}}

@INPROCEEDINGS{GAN-based,
  author={Arisoy, Sertac and Nasrabadi, Nasser M. and Kayabol, Koray},
  booktitle={2020 28th European Signal Processing Conference (EUSIPCO)}, 
  title={GAN-based Hyperspectral Anomaly Detection}, 
  year={2021},
  volume={},
  number={},
  pages={1891-1895},
  doi={10.23919/Eusipco47968.2020.9287675}}

@ARTICLE{Transformer-based,
  author={Xiao, Song and Zhang, Tian and Xu, Zhangchun and Qu, Jiahui and Hou, Shaoxiong and Dong, Wenqian},
  journal={IEEE Journal of Selected Topics in Applied Earth Observations and Remote Sensing}, 
  title={Anomaly Detection of Hyperspectral Images Based on Transformer With Spatial–Spectral Dual-Window Mask}, 
  year={2023},
  volume={16},
  number={},
  pages={1414-1426},
  doi={10.1109/JSTARS.2022.3232762}}

@ARTICLE{manifold-AE,
  author={Lu, Xiaoqiang and Zhang, Wuxia and Huang, Ju},
  journal={IEEE Transactions on Geoscience and Remote Sensing}, 
  title={Exploiting Embedding Manifold of Autoencoders for Hyperspectral Anomaly Detection}, 
  year={2020},
  volume={58},
  number={3},
  pages={1527-1537},
  doi={10.1109/TGRS.2019.2944419}}

@ARTICLE{chessboad-topology,
  author={Gao, Lianru and Sun, Xiaotong and Sun, Xu and Zhuang, Lina and Du, Qian and Zhang, Bing},
  journal={IEEE Transactions on Geoscience and Remote Sensing}, 
  title={Hyperspectral Anomaly Detection Based on Chessboard Topology}, 
  year={2023},
  volume={61},
  number={},
  pages={1-16},
  doi={10.1109/TGRS.2023.3249748}}

@inproceedings{topology-based,
author = {Bill Basener and Emmett J. Ientilucci and David W. Messinger},
title = {Anomaly detection using topology},
volume = {6565},
booktitle = {Algorithms and Technologies for Multispectral, Hyperspectral, and Ultraspectral Imagery XIII},
editor = {Sylvia S. Shen and Paul E. Lewis},
organization = {International Society for Optics and Photonics},
publisher = {SPIE},
pages = {65650J},
year = {2007},
doi = {10.1117/12.745429},
URL = {https://doi.org/10.1117/12.745429}
}

@ARTICLE{IGPAD,
  author={Sheng, Jiahui and Li, Xiaorun and Chen, Shuhan},
  journal={IEEE Transactions on Geoscience and Remote Sensing}, 
  title={Iterative Graph Propagation for Hyperspectral Anomaly Detection}, 
  year={2024},
  volume={62},
  number={},
  pages={1-14},
  doi={10.1109/TGRS.2024.3430111}}

@INPROCEEDINGS{graph-based-1,
  author={Messinger, D. W. and Albano, J.},
  booktitle={2011 3rd Workshop on Hyperspectral Image and Signal Processing: Evolution in Remote Sensing (WHISPERS)}, 
  title={A graph theoretic approach to anomaly detection in hyperspectral imagery}, 
  year={2011},
  volume={},
  number={},
  pages={1-4},
  doi={10.1109/WHISPERS.2011.6080899}}

@article{graph-based-2,
  title={Graph Laplacian for image anomaly detection},
  author={Verdoja, Francesco and Grangetto, Marco},
  journal={Machine Vision and Applications},
  volume={31},
  number={1},
  pages={11},
  year={2020},
  publisher={Springer}
}

@ARTICLE{ELRSF-SP,
  author={Xiao, Qingjiang and Zhao, Liaoying and Chen, Shuhan and Li, Xiaorun},
  journal={IEEE Journal of Selected Topics in Applied Earth Observations and Remote Sensing}, 
  title={Hyperspectral Anomaly Detection via Enhanced Low-Rank and Smoothness Fusion Regularization Plus Saliency Prior}, 
  year={2024},
  volume={17},
  number={},
  pages={18987-19002},
  doi={10.1109/JSTARS.2024.3478848}}

@ARTICLE{GRX,
  author={Reed, I.S. and Yu, X.},
  journal={IEEE Transactions on Acoustics, Speech, and Signal Processing}, 
  title={Adaptive multiple-band CFAR detection of an optical pattern with unknown spectral distribution}, 
  year={1990},
  volume={38},
  number={10},
  pages={1760-1770},
  doi={10.1109/29.60107}}

@ARTICLE{CRD,
  author={Li, Wei and Du, Qian},
  journal={IEEE Transactions on Geoscience and Remote Sensing}, 
  title={Collaborative Representation for Hyperspectral Anomaly Detection}, 
  year={2015},
  volume={53},
  number={3},
  pages={1463-1474},
  doi={10.1109/TGRS.2014.2343955}}

@ARTICLE{TVSDM,
  author={Cheng, Tongkai and Wang, Bin},
  journal={IEEE Transactions on Geoscience and Remote Sensing}, 
  title={Total Variation and Sparsity Regularized Decomposition Model With Union Dictionary for Hyperspectral Anomaly Detection}, 
  year={2021},
  volume={59},
  number={2},
  pages={1472-1486},
  doi={10.1109/TGRS.2020.3004478}}

@ARTICLE{GTVLRR,
  author={{T. Cheng and B. Wang}},
  journal={IEEE Transactions on Geoscience and Remote Sensing}, 
  title={Graph and Total Variation Regularized Low-Rank Representation for Hyperspectral Anomaly Detection}, 
  year={2020},
  volume={58},
  number={1},
  pages={391-406},
  }

@ARTICLE{LSMAD,
  author={Zhang, Yuxiang and Du, Bo and Zhang, Liangpei and Wang, Shugen},
  journal={IEEE Transactions on Geoscience and Remote Sensing}, 
  title={A Low-Rank and Sparse Matrix Decomposition-Based Mahalanobis Distance Method for Hyperspectral Anomaly Detection}, 
  year={2016},
  volume={54},
  number={3},
  pages={1376-1389},
  doi={10.1109/TGRS.2015.2479299}}

@ARTICLE{PTA,
  author={Li, Lu and Li, Wei and Qu, Ying and Zhao, Chunhui and Tao, Ran and Du, Qian},
  journal={IEEE Transactions on Neural Networks and Learning Systems}, 
  title={Prior-Based Tensor Approximation for Anomaly Detection in Hyperspectral Imagery}, 
  year={2022},
  volume={33},
  number={3},
  pages={1037-1050},
  doi={10.1109/TNNLS.2020.3038659}}

@ARTICLE{RGAE,
  author={Fan, Ganghui and Ma, Yong and Mei, Xiaoguang and Fan, Fan and Huang, Jun and Ma, Jiayi},
  journal={IEEE Transactions on Geoscience and Remote Sensing}, 
  title={Hyperspectral Anomaly Detection With Robust Graph Autoencoders}, 
  year={2022},
  volume={60},
  number={},
  pages={1-14},
  doi={10.1109/TGRS.2021.3097097}}

@ARTICLE{Auto-AD,
  author={Wang, Shaoyu and Wang, Xinyu and Zhang, Liangpei and Zhong, Yanfei},
  journal={IEEE Transactions on Geoscience and Remote Sensing}, 
  title={Auto-AD: Autonomous Hyperspectral Anomaly Detection Network Based on Fully Convolutional Autoencoder}, 
  year={2022},
  volume={60},
  number={},
  pages={1-14},
  doi={10.1109/TGRS.2021.3057721}}

@ARTICLE{3D-ROC-TD,
  author={Chang, Chein-I},
  journal={IEEE Transactions on Geoscience and Remote Sensing}, 
  title={An Effective Evaluation Tool for Hyperspectral Target Detection: 3D Receiver Operating Characteristic Curve Analysis}, 
  year={2021},
  volume={59},
  number={6},
  pages={5131-5153},
  doi={10.1109/TGRS.2020.3021671}}

@ARTICLE{3D-ROC-AD,
  author={{C.-I. Chang}},
  journal={IEEE Transactions on Geoscience and Remote Sensing}, 
  title={Comprehensive Analysis of Receiver Operating Characteristic (ROC) Curves for Hyperspectral Anomaly Detection}, 
  year={2022},
  volume={60},
  number={},
  pages={1-24},
  doi={10.1109/TGRS.2022.3211786}}

\end{document}